\documentclass[sn-mathphys,Numbered]{sn-jnl}

\usepackage{graphicx}%
\usepackage{multirow}%
\usepackage{amsmath,amssymb,amsfonts}%
\usepackage{amsthm}%
\usepackage{mathrsfs}%
\usepackage[title]{appendix}%
\usepackage{xcolor}%
\usepackage{textcomp}%
\usepackage{manyfoot}%
\usepackage{booktabs}%
\usepackage{algorithm}%
\usepackage{algorithmicx}%
\usepackage{algpseudocode}%
\usepackage{listings}%
\usepackage{lettrine}
\usepackage{pifont}
\usepackage{changes}
\usepackage{tabularx}
\usepackage{booktabs} 
\usepackage{graphicx}
\usepackage{subcaption}
\usepackage{colortbl}
\definecolor{lightblue}{rgb}{0.68, 0.85, 0.9}

\DeclareUnicodeCharacter{2212}{-}

\theoremstyle{thmstyleone}%
\theoremstyle{thmstyletwo}%

\theoremstyle{thmstylethree}%

\begin{document}

\definechangesauthor[color=blue]{R1}
\definechangesauthor[color=purple]{R2}
\definechangesauthor[color=orange]{R3}

\title[Article Title]{SWAG: Long-term Surgical Workflow Prediction with Generative-based Anticipation}


\author*[]{\fnm{Maxence} \sur{Boels}*}\email{maxence.boels@kcl.ac.uk}
\author[]{\fnm{Yang} \sur{Liu}}
\author[]{\fnm{Prokar} \sur{Dasgupta}}

\author{\fnm{Alejandro} \sur{Granados}}

\author{\fnm{Sebastien} \sur{Ourselin}}

\affil[]{\orgdiv{Surgical and Interventional Engineering}, \orgname{King's College London}}

\abstract{
\textbf{Purpose}
While existing approaches excel at recognising current surgical phases, they provide limited foresight and intraoperative guidance into future procedural steps. Similarly, current anticipation methods are constrained to predicting short-term and single events, neglecting the dense, repetitive, and long sequential nature of surgical workflows. To address these needs and limitations, we propose SWAG (Surgical Workflow Anticipative Generation), a framework that combines phase recognition and anticipation using a generative approach.

\textbf{Methods}
This paper investigates two distinct decoding methods—single-pass (SP) and auto-regressive (AR)—to generate sequences of future surgical phases at minute intervals over long horizons. We propose a novel embedding approach using class transition probabilities to enhance the accuracy of phase anticipation. Additionally, we propose a generative framework using remaining time regression to classification (R2C). SWAG was evaluated on two publicly available datasets, Cholec80 and AutoLaparo21.

\textbf{Results}
Our single-pass model with class transition probability embeddings (SP*) achieves 32.1\% and 41.3\% F1 scores over 20 and 30 minutes on Cholec80 and AutoLaparo21, respectively. Moreover, our approach competes with existing methods on phase remaining time regression, achieving weighted mean absolute errors of 0.32 and 0.48 minutes for 2- and 3-minute horizons.

\textbf{Conclusion}
SWAG demonstrates versatility across generative decoding frameworks and classification and regression tasks to create temporal continuity between surgical workflow recognition and anticipation. Our method provides steps towards intraoperative surgical workflow generation for anticipation. Project: \url{https://maxboels.github.io/swag}.
}


\maketitle

\section{Introduction}\label{Introduction}

\lettrine[lraise=0.25, nindent=0em, slope=-.5em]{S}{urgical workflow anticipation} has the potential to enhance operating room efficiency and patient safety by predicting future surgical events during procedures. Anticipation allows for better preparation, reduces cognitive loads, and improves coordination among surgical teams~\cite{Sexton148, Yurko2010}. While preoperative planning provides an initial framework, it often falls short in addressing the dynamic environment and decision-making required during surgery.

 Current approaches predominantly focus on surgical phase recognition, identifying the present phase, step, or action~\cite{czempiel2020tecno, liu2023lovit, liu2023skit}. These methods are valuable for postoperative analysis but offer limited assistance in real-time intraoperative decision-making. They cannot anticipate future events to allow for dynamic planning adjustments, which could potentially improve surgical outcomes.
Remaining Surgery Duration (RSD) estimation has been actively researched for over two decades~\cite{aksamentov2017deep, twinanda2018rsdnet, rivoir2019unsupervised, marafioti2021catanet, wu2023bdnet, wijekoon2024pitrsdnet}, demonstrating clinical interest in long-term predictions—i.e. predicting the remaining time until the end of the surgery. Other studies explored surgical workflow anticipation, such as predicting the next phases and instrument occurrences~\cite{rivoir2020rethinking, YUAN2022102611}. However, these methods are limited by their requirement to predict a single event per class, disregarding scenarios where multiple future occurrences may exist within a fixed time horizon.

Generative models, such as GPT models~\cite{radford2019language}, address this challenge by generating sequences of tokens that can vary in both length and class frequency. In surgical workflow anticipation, auto-regressive (AR)~\cite{radford2019language} and single-pass (SP) decoding models~\cite{wang2023memoryandanticipation} demonstrate the potential to predict continuous sequences of surgical actions. Unlike conventional remaining time regression approaches, generative models output multiple tokens, extending the observed sequence of events to a plausible future representation of the surgical workflow.

In this work, we present SWAG (Surgical Workflow Anticipative Generation), a generative model designed to unify phase recognition and anticipation in surgical workflows. Our main contributions are as follows:

\begin{enumerate} 
    \item We introduce SWAG, a generative model that combines surgical phase recognition and anticipation, for long-term sequential future prediction of the workflow.
    \item We provide an extensive comparison of two generative decoding methods—single-pass and auto-regressive—alongside two tasks—classification and regression—to generate continuous sequences of present and future surgical phases.
    \item We propose a novel prior knowledge embedding method using future class transition probabilities improving predictive performances. 
    \item We conduct a comprehensive evaluation on the Cholec80 and AutoLaparo21 datasets, demonstrating SWAG’s capabilities on different surgical procedures.
\end{enumerate}

\section{Related Work}\label{sec2}

\noindent{\textbf{Surgical Phase Recognition.}} Early research on surgical phase recognition relied on probabilistic graphical models with instrument usage~\cite{DBLP:conf/miccai/BlumFN10}. Later, convolutional neural networks were used to learn spatial representations from surgical video frames, while temporal relations among video frames were captured with recurrent neural networks~\cite{DBLP:phd/hal/Twinanda17}.
Temporal convolutional networks were later able to increase the receptive field~\cite{czempiel2020tecno}. Recent works have successfully used Transformers~\cite{vaswani2017attention, dosovitskiy2020image} to capture critical information between frames using short attention windows like in Trans-SVNet~\cite{gao2021transsvnet} and LoViT~\cite{liu2023lovit}, whereas SKiT~\cite{liu2023skit} proposed an efficient long-term compression approach achieving state-of-the-art performance on online surgical phase recognition. SAHC~\cite{ding2022exploring} proposes to model surgical workflows at both frame and segment levels to better capture phase transitions. Limitations between anticipation and batch normalisation for surgical workflow analysis are presented in BNPitfalls~\cite{rivoir2024pitfalls}. SurgFormer~\cite{yang2024surgformer} introduces a transformer architecture with hierarchical temporal attention for phase recognition. Although these approaches offer substantial advancements for postoperative video analysis, we focus on intraoperative decision-making by adding anticipative capabilities to the recognition system. Anticipating the surgical workflow over long time horizons could enhance real-time decision support by forecasting upcoming surgical phases, enabling smoother, more responsive guidance during procedures.

\noindent{\textbf{Surgical Workflow Anticipation.}} Most anticipation approaches in surgery have explored surgical workflow anticipation by predicting the remaining time until the end of surgery \cite{aksamentov2017deep, twinanda2018rsdnet, rivoir2019unsupervised, marafioti2021catanet, wu2023bdnet, wijekoon2024pitrsdnet}, next instruments~\cite{rivoir2020rethinking} or next phases occurrence~\cite{YUAN2022102611, ban2021suprgan, boels2024supra}. Bayesian~\cite{rivoir2020rethinking} was proposed to anticipate tool usage which was then used as a baseline in IIA-Net~\cite{YUAN2022102611} for surgical phase anticipation. IIA-Net~\cite{YUAN2022102611} was then introduced to leverage instrument interaction for next-phase occurrence regression. Although IIA-Net requires pre-trained models for tool detection and segmentation, we compare our method to this approach and their implementation of Bayesian~\cite{rivoir2020rethinking}. Both methods were formulated as a regression problem, for surgical phase anticipation. SUPR-GAN~\cite{ban2021suprgan} uses LSTMs within an encoder-decoder architecture to predict surgical phases over 15 seconds. A discriminator is used to train the model, using a generative adversarial network (GAN) approach. In contrast, our SWAG model addresses long-term surgical workflow anticipation, predicting sequences up to 60 minutes while unifying recognition and anticipation tasks. Zhang et al. \cite{zhang2022towards} proposes a graph network with bounding boxes as inputs to predict the occurrence of instruments or phases within 2-, 3-, and 5-minute horizons. Later, Hypergraph-Transformer (HGT) was proposed in ~\cite{yin2024hypergraphtransformer} to detect and predict action triplets over 4 seconds. Other methods were also proposed for gesture anticipation as low-level motion planning~\cite{ginesi2020autonomous}, instrument trajectory prediction~\cite{qin2020davincinet, zhang2023laparoscopic}. Most methods focus on the next occurrence regression, failing to provide a comprehensive view of the entire surgical workflow, leaving a blind spot for events beyond the first predicted occurrence. Unlike previous works, SWAG uses a generative approach for future phase classification. Our approach addresses some gaps and limitations by predicting sequences of arbitrary length and frequency, and unifying recognition and anticipation tasks.

\section{Methods}\label{Methods}

In this section, we present our proposed method for jointly addressing surgical phase recognition and anticipation. Our model, SWAG, is designed to predict the current phase while anticipating the occurrence of future phases over long-time horizons.

\subsection{Task Formulation}

Our primary task is to predict the current and future surgical phases over a horizon of \( N = h_N \) minutes, conditioned on observed frames \( X_t = \{x_0,\ldots,x_t\} \), where \( x_i \in \mathcal{X}_T \) represents all video frames from the surgical procedure. We define the mapping function as follows:
\begin{equation}
    f_{\theta}: X_{\leq t} \mapsto \bigl(y_{t+h_0\cdot60}, \ldots, y_{t+h_N\cdot60}\bigr)
\end{equation}
where \( X_{\leq t} \) denotes the sequence of observed frames up to time \( t \). The anticipation sequence \( \{h_n\}_{n=1}^{N} \) is a predefined set of time steps, where each \( h_n \) corresponds to the number of minutes into the future for which we predict the surgical phase. The sequence is defined as:  
\begin{equation}
    h_0 = 0, \quad h_1, h_2, \dots, h_N \in \{1, 2, \dots, N\}
\end{equation}
where \( h_0 \) represents the current phase prediction and \( h_N \) defines the maximum anticipation horizon.

For the time regression task, we predict the remaining time until each phase's next occurrence within a fixed horizon of \([0, N]\) minutes, where \( 0 \) indicates the phase is currently active, and \( N \) means it will not occur within the horizon. Our setup follows \cite{rivoir2020rethinking,YUAN2022102611}. We predict phases every 60 seconds based on our ablations (see Fig.~4 in SM).

\begin{figure}[h]
    \centering
    \includegraphics[width=1.0\textwidth]{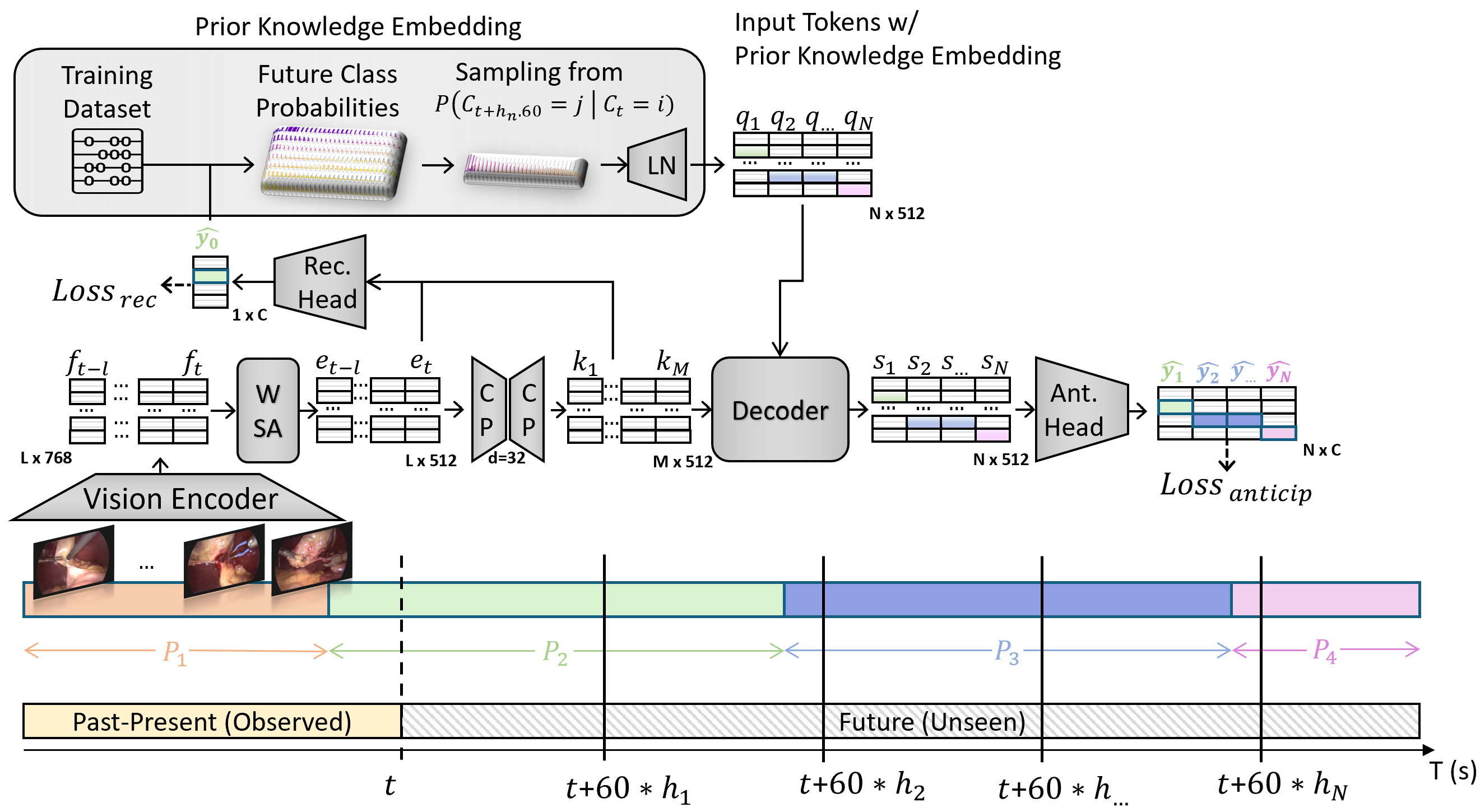}
    \caption{Overview of the SWAG-SP* model architecture. The model processes surgical video data in two main paths: (1) A recognition path with $L=24*60$ frame embeddings and $M=24$ compressed frame representations are encoded into $D=512$-dimensional embeddings through a Vision Encoder, a Windowed Self-Attention (WSA) encoder and a Compression and max-Pooling (CP) bottleneck to recognise the current surgical phase with a classification head (Loss$_{recognition}$). (2) A generative path samples $N$ times from future class probabilities extracted from the training set. These probabilities are combined with temporal position embeddings to form input tokens $Q_t=\{{q_1, q_2, ..., q_N}\}$, and passed to a Decoder that conditions the compressed frame embeddings i.e. context tokens $K_t=\{{k_1, k_2, ..., k_M}\}$ to predict $N$ future surgical phases at $60$ seconds intervals (Loss$_{anticipation}$). The timeline shows the model operating on past-present observed frames from phases P$_1$ (orange) and P$_2$ (green) and predicting probabilities $N \times C$ of future phases.}
    \label{fig:swag_sp_star}
\end{figure}

\subsection{SWAG Model Architecture}

Our proposed model, illustrated in Figure~\ref{fig:swag_sp_star}, includes a Vision Encoder followed by a Windowed Self-Attention encoder (WSA), then a linear Compression and max-Pooling block (CP)~\cite{liu2023skit}, and finally a future Decoder module that can operate either in a single-pass (SP) or an auto-regressive (AR) approach. While our model uses classification for the phase recognition task, it can use classification or regression for phase anticipation. This architecture also includes a novel Prior Knowledge Embedding relying on the predicted current class $\hat{y}_{t+h_0\cdot60}$ and future indices from $h_1$ to $h_N$.

\noindent\textbf{Vision Encoder.} Similar to LoViT, we fine-tune a pre-trained ViT~\cite{dosovitskiy2020image} using the AVT~\cite{girdhar2021anticipative} approach, which trains on short segments of consecutive frames. This procedure allows the model to incorporate limited temporal context while learning spatial features. At inference, we then extract a single $768$-dimensional embedding per frame $f_t$, serving as the initial feature representation.

\noindent\textbf{Window Self-Attention (WSA).} To recognise the observed frame at time $t$, we use the extracted features as input sequence $F_t = {\{f_{t-L}, \ldots, f_t}\}$ with a length of $L=1440$ and $D=768$ dimensions. We then use a sliding window of width $W=20$ to perform self-attention over the extracted image features with no overlap between windows and get an output sequence of the same length and dimensionality, $E_t = {\{e_{t-L}, \ldots, e_t}\}$.

\noindent\textbf{Compression and Pooling (CP)}. We implement two temporal pooling strategies: global key-pooling~\cite{liu2023skit} and interval-pooling, which are used differently in our single-pass and auto-regressive decoders. The single-pass method exclusively uses global key-pooling, while the auto-regressive approach leverages both methods. Specifically, the recognition branch in both models uses global key-pooling. For global key-pooling, the temporal features $E_t = {e_{t-L}, \ldots, e_t}$ are first projected into a lower-dimensional latent space ($d=32$) as $E'_t$ using a linear layer, then compressed via a cumulative max-pooling approach into $M$ tokens $K_t = {k_{1}, \ldots, k_M}$. Following~\cite{liu2023skit}, this involves first computing a single pooled representation of the compressed features:
$p = \max{\{e'_{t-L}, \ldots, e'_{t-M}\}}$
Then, for each of the most recent $M$ frames, we compute a cumulative maximum:
$k_m = \max{\{p, e'_{t-M+1}, \ldots, e'_{t-M+m}\}}$ for $m = 1, 2, \ldots, M$
This creates tokens that progressively incorporate more recent information while maintaining context from all previous frames. The auto-regressive approach, trained with causal masking on input sequences, requires temporal consistency between the training sequence and the generated outputs tokens during inference. Therefore, we developed \textit{interval-pooling} to aggregate spatial information at regular temporal intervals. Interval-pooling performs max-pooling over consecutive 60-second intervals, with a fixed start time at $t-L$. For each new token, the right-side index extends by $60$ seconds, effectively compressing $60$ frame embeddings into a single token. This yields $M$ context tokens, where $M$ corresponds to the sequence length divided by the 60-second interval (e.g., compressing 24 minutes into 24 tokens: $\frac{1440\text{ seconds}}{60\text{ seconds}}$). This approach ensures a progressive and cumulative aggregation of information which is consistent with the time interval in the generated sequence during inference.

\begin{figure}[t]%
    \centering
    \includegraphics[width=1.0\textwidth]{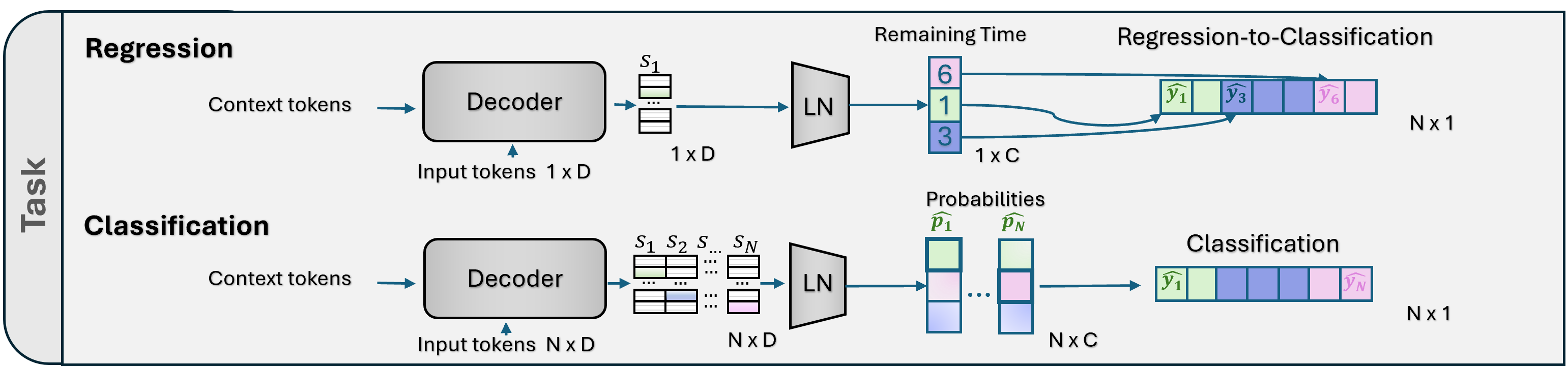}
    \caption{Single-Pass regression and classification tasks. The regression task (top) uses a single $1 \times D$ input token to predict the remaining time before each next class occurrence, which can be transformed into a classification task using our regression-to-classification (R2C) method. The direct classification task (bottom) uses $N$ input tokens to predict $N$ minutes in the future.}
    \label{fig:swag_tasks}
\end{figure}

\noindent\textbf{Decoder}. Our two decoding approaches (see Fig.~\ref{fig:swag_tasks}) can be described as follows:

1) \textbf{SWAG-SP (Single-Pass)}: The vanilla transformer decoder receives $N$ input tokens $Q_t=\{{q_1, ..., q_N}\}$ and generates $N$ output tokens $S_t=\{{s_1, ..., s_N}\}$ in a single forward pass, each token represents $60$ seconds over $N$ minutes. During training, we use a special teacher-forcing approach, using the ground-truth current phase $y_t$ to sample from the correct conditional probability distribution  $P(y_{t+h_n\cdot60} = j \mid y_t = i)$ at future minute index $h_n$, $N$ times. The $Q_t$ inputs are initialized with prior knowledge corresponding to conditional future class probabilities extracted from the training set (see Sec. 1.2 in SM) and sinusoidal positional encoding. This generative approach is conditioned using cross-attention between the context tokens $K_t = {\{k_{1}, \ldots, k_M}\}$ and the $Q_t$ inputs to generate all future tokens $S_t$ in a single forward pass, allowing for efficient parallel computing and reducing inference time.

2) \textbf{SWAG-AR (Auto-Regressive)}: We use the GPT-2 model as our auto-regressive decoder~\cite{radford2019language} which uses causal masking on the features extracted with interval-pooling to predict the next token in the sequence. Teacher forcing is not used as our approach uses the learned frame embeddings from the recognition task rather than embedding ground-truth labels as inputs. During inference, the decoder iteratively uses past predictions as inputs until its generate $N=h_N$ future tokens as its anticipated horizon.

\noindent\textbf{Recognition Head (Rec. Head)}. We fuse the key-pooled features $K_t$ with the windowed self-attention features $E_t=\{{e_{t-M-1}, ..., e_t}\}$ using a skip connection~\cite{liu2023skit}. The resulting fused feature vectors are then passed through a classification layer to predict $M$ class probabilities $\{\hat{p_1}, \ldots, \hat{p_M}\}$, which estimate the class labels assigned to the input frames $\{{x_{t-M-1}, \ldots, x_t}\}$.

\noindent\textbf{Anticipation Head (Ant. Head)}. We use a linear layer (LN) and softmax to predict future phase probabilities $\{\hat{p_1}, \ldots, \hat{p_N}\}$ with shape $N \times C$.

Note that, the regression task outputs $1 \times C$ values for the remaining time until the next phases, including the End-Of-Surgery (EOS) class. While remaining time regression could be performed via auto-regressive decoding with a single iteration, we opt for a single-pass decoder since a single token can represent specific transitions over time, and also due to the stronger performance on the classification task. Nevertheless, future work could explore generating multiple transitions per class (e.g., one for each generated token) to narrow potential performance gaps relative to an SP scheme.

\noindent\textbf{SWAG-SP* (with Prior Knowledge Embedding).} Our extracted prior knowledge captures the conditional probabilities of phase transitions at specific future times and is exclusively derived from the training set. Specifically, we compute the probability of having a future class $j$ at time step $t+h_n\cdot60$ given the current observed class $i$, denoted as \( P(y_{t+h_n\cdot60} = j \mid y_t = i) \). These probabilities form a tensor \( P \), where each entry \( P_{i,j,h_n\cdot60} \) represents the likelihood of transitioning between classes at each anticipated time. We use these probabilities to initialise future token embeddings, assigning each future token at \( t+h_n\cdot60 \) the probability vector \( P_{i,:,h_n\cdot60} \).

\noindent\textbf{SWAG-SP-R2C (Regression-to-Classification).} We demonstrate a possible mapping from regression outputs to classification via our regression-to-classification (R2C) method, which converts continuous time predictions into discrete phase intervals by sorting and binning predicted times. This approach can be used if one wishes to derive a phase sequence from a model originally trained for regression. Specifically, remaining time predictions for each class are mapped to a discrete sequence of $N$ bins containing the predicted class integers $\hat{Y}_t=\{\hat{y}_0, \hat{y}_1, \ldots, \hat{y}_N\}$, in ascending order of occurrence. We use the current predicted class $\hat{y}_0$ to fill the sequence up to the first next class occurrence and use cross-entropy and mean squared error losses for the classification and regression tasks, respectively.

\subsection{Experimental Setup}

\noindent\textbf{Datasets.} We evaluate our work on two publicly available surgical phase datasets: \textit{Cholec80} (C80) \cite{twinanda2017endonet} and \textit{AutoLaparo21} (AL21) \cite{wang2022autolaparo}. For Cholec80, we adopt the same dataset splits as in \cite{funke2025tunes}, using 32 videos for training, 8 for validation, and 40 for testing. Similarly, for AL21, we apply the 10/4/7 split for training, validation, and testing, following \cite{wang2022autolaparo, liu2023skit}. For the regression task, we follow a 60/20 train-test split as in \cite{rivoir2020rethinking, YUAN2022102611}. Both datasets were sampled at 1 frame per second (fps) following previous works \cite{gao2021transsvnet, liu2023lovit}. Across both datasets, we train on the 7 surgical phases, including an additional end-of-surgery (EOS) class, while disregarding tool annotations.

\noindent\textbf{Training Details.} We train our model on a maximal horizon of $N=h_N$ minutes and evaluate it at intermediate steps without re-training. We use a simple weighted cross-entropy loss for the classification task and mean squared error loss for the regression task. Additional details on the training procedure are provided in the supplementary material (Sec. 1.1 in SM).

\noindent\textbf{Anticipation Time.} Our method evaluates phase anticipation across multiple time horizons to assess both short-term and long-term predictive capabilities. This approach is justified by the substantial duration of procedures in our datasets: C80 training procedures average $38$ minutes, while AL21 training videos average $66$ minutes. By supervising our models on extended time horizons that encompass most of these workflows, we gain an important practical benefit: the ability to estimate the completion time of the entire surgical procedure from its earliest stages.

\noindent\textbf{Classification Task.} Since long-term future phase classification has not been studied in previous work, direct comparison to other surgical workflow anticipation methods is not possible. Therefore, we assess our generative approaches against two baseline methods: a simple continuation model (Naive1) and a conditional-probabilistic model (Naive2).
\textbf{Naive1} baseline extends the current recognised class over the future predicted horizon for $N$ minutes. While effective for short-term predictions, its performance naturally degrades over time with new class transitions.
\textbf{Naive2} baseline predicts future phases by sampling from the class probability distribution \( P_{i,j,h_n\cdot60} \), conditioned on the predicted current class \( i \) and at the anticipated time index \( h_n \). This baseline is effective due to the strong priors inherent in structured workflows, leveraging high recognition accuracy and precisely defined anticipation times.

\noindent\textbf{Regression Task.} We include a secondary evaluation to compare our method to previous works. Specifically, we benchmark against Bayesian~\cite{rivoir2020rethinking} and IIA-Net~\cite{YUAN2022102611}, both predicting the remaining time until the next phase occurrence. IIA-Net relies on instrument presence and phase labels for supervision, while we solely use the latter. We did not include results from the Trans-SVNet~\cite{jin2022trans} method as their evaluation was limited to a single horizon (5 minutes) and did not report the weighted MAE score (see Sec. \ref{sec:eval-metrics}). We extend the phase anticipation regression task to long horizons to include the remaining surgery duration (RSD) evaluation following previous works \cite{twinanda2018rsdnet, marafioti2021catanet, wu2023bdnet}.

\subsection{Evaluation Metrics}\label{sec:eval-metrics}
We use a combination of classification and regression metrics to evaluate our methodology. These metrics are calculated for each anticipation time, allowing us to analyse the model's performance through time. For classification, we use weighted F1 scores which account for class imbalance by assigning importance proportional to class frequency. Surgical workflows are naturally imbalanced due to the inherent structure of procedural steps. To limit EOS class dominance, we cap EOS samples at 4 minutes for Cholec80 and 8 minutes for AutoLaparo21. We also introduce SegF1, a segment-based F1 score that evaluates temporal coherence by matching predicted segments to ground truth using IoU thresholding, penalizing oversegmentation while rewarding correct phase boundaries. Detailed calculation is provided in the Supplementary Material.
For regression tasks, we use the Mean Absolute Error (MAE) to evaluate the model's performance on predicting the remaining time until the next phase occurrence. This metric quantifies the average absolute prediction errors in minutes. We use the \textit{w}MAE, \textit{in}MAE, and \textit{out}MAE for \textit{weighting} samples \textit{outside} and \textit{inside} the temporal horizon~\cite{rivoir2020rethinking, YUAN2022102611}.
Model selection is performed separately for each dataset using the validation set. For classification, we select the model with the highest overall mean weighted F1 score. For regression, we choose the model with the lowest weighted MAE.

\section{Results}
\subsection{Surgical Phase Anticipation}
\begin{figure}[h]
    \centering \includegraphics[width=1.\textwidth]{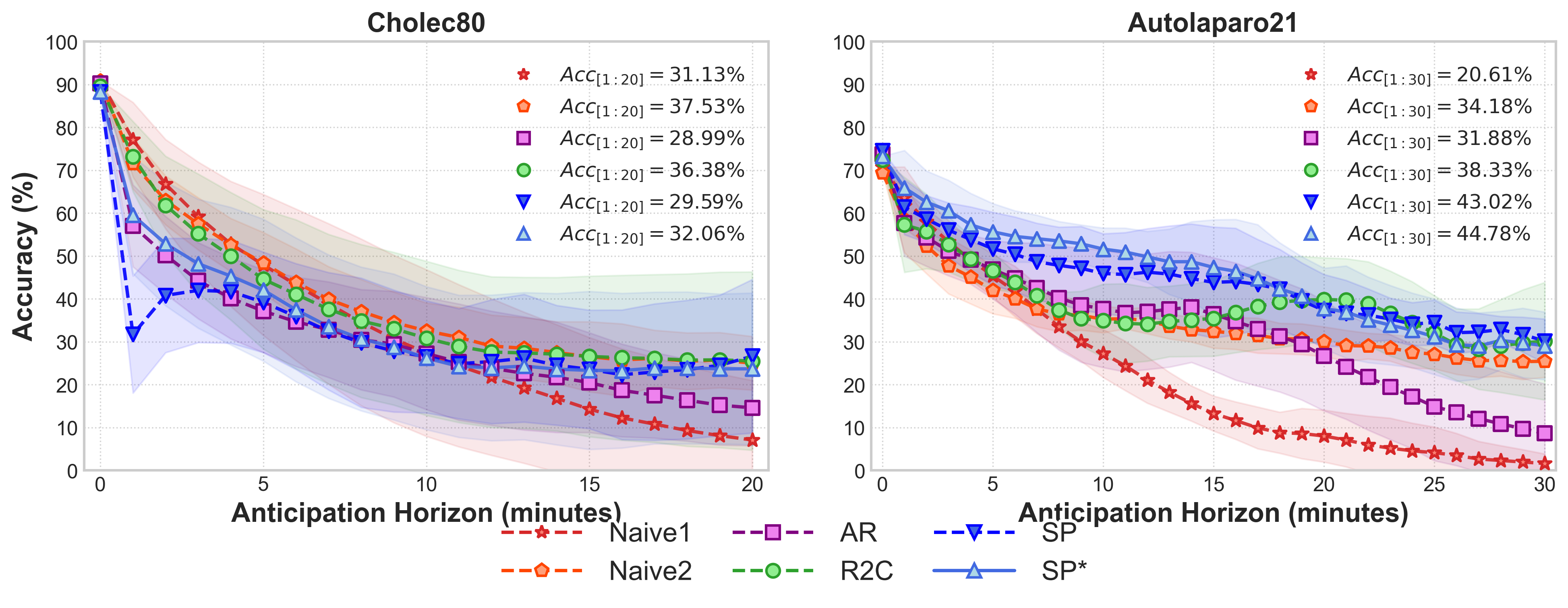}
    \caption{Models performance for surgical phase recognition and anticipation on Cholec80 and AutoLaparo21. Accuracy ($Acc_{[1:h_n]}$) over $h_n$ minutes (top right corners).}
    \label{fig:f1_scores_over_time}
\end{figure}

Table~\ref{tab:main_results} compares our models against two naive baselines. While Naive2 achieves strong anticipation performance on Cholec80 (F1 = 39.5\%), our SP* model excels on the more challenging AutoLaparo21 dataset (F1 = 41.3\%), demonstrating superior generalisation to complex, variable procedures. Notably, despite Naive2's strong minute-level results, it performs poorly on multiple minutes intervals due to discontinuous predictions, whereas our approach maintains consistent segment-level (SegF1) performance. Figure~\ref{fig:f1_scores_over_time} illustrates performance degradation across anticipation horizons.

\begin{table*}[t]
\centering
\small
\caption{\textbf{Surgical Phase Recognition, Anticipation, and Segment-level Performance.} Frame-level metrics use weighted averaging to handle class imbalance. Recognition: accuracy/F1-score at current time ($t=0$). Anticipation: mean F1-score over horizon and IoU-based segment F1.
}
\begin{tabular}{l|cccc|cccc}
\toprule
\multirow{3}{*}{\textbf{Methods}} & \multicolumn{4}{c|}{\textbf{Cholec80}} & \multicolumn{4}{c}{\textbf{AutoLaparo21}} \\
 & \multicolumn{2}{c}{\textbf{Recognition}} & \multicolumn{2}{c}{\textbf{Anticipation}} & \multicolumn{2}{c}{\textbf{Recognition}} & \multicolumn{2}{c}{\textbf{Anticipation}} \\
 & \textbf{Acc} & \textbf{F1} & \textbf{F1} & \textbf{SegF1} & \textbf{Acc} & \textbf{F1} & \textbf{F1} & \textbf{SegF1} \\
\midrule
Naive1$^\dagger$ & \cellcolor{lightblue!25}\textbf{90.8} & \underline{91.1} & 29.3 & 24.5 & 70.7 & 71.0 & 19.6 & 16.4 \\
Naive2$^\dagger$ & \underline{90.8} & \cellcolor{lightblue!25}\textbf{91.2} & \cellcolor{lightblue!25}\textbf{39.5} & 11.9 & 69.4 & 69.3 & 34.3 & 10.7 \\
\midrule
AR & 90.3 & 90.7 & 27.8 & 25.0 & \underline{73.7} & \underline{72.9} & 29.3 & 23.3 \\
R2C & 89.6 & 89.8 & \underline{36.1} & \cellcolor{lightblue!25}\textbf{32.5} & 72.4 & 71.8 & 32.9 & 29.2 \\
SP & 88.3 & 88.8 & 29.4 & 23.8 & \cellcolor{lightblue!25}\textbf{74.6} & \cellcolor{lightblue!25}\textbf{73.5} & \underline{38.4} & \underline{30.8} \\
SP$^*$ & 88.3 & 88.8 & 32.1 & \underline{29.8} & 73.3 & 72.7 & \cellcolor{lightblue!25}\textbf{41.3} & \cellcolor{lightblue!25}\textbf{34.8} \\
\bottomrule
\end{tabular}
\begin{tablenotes}
\footnotesize
\item \textbf{Note:} Frame-level metrics use weighted averaging for class imbalance. Recognition: Acc/F1 at $t=0$. Anticipation: mean F1 over horizon. Segment: IoU-based F1 with EOS weighting (max 4/8 EOS samples per sequence). \cellcolor{lightblue!25}\textbf{Best} performance highlighted, \underline{second-best} underlined. $^\dagger$ Baseline models (recognition only). $^*$ Prior knowledge initialization.
\end{tablenotes}
\label{tab:main_results}
\end{table*}

\begin{figure}[t]
    \includegraphics[width=1.0\linewidth]{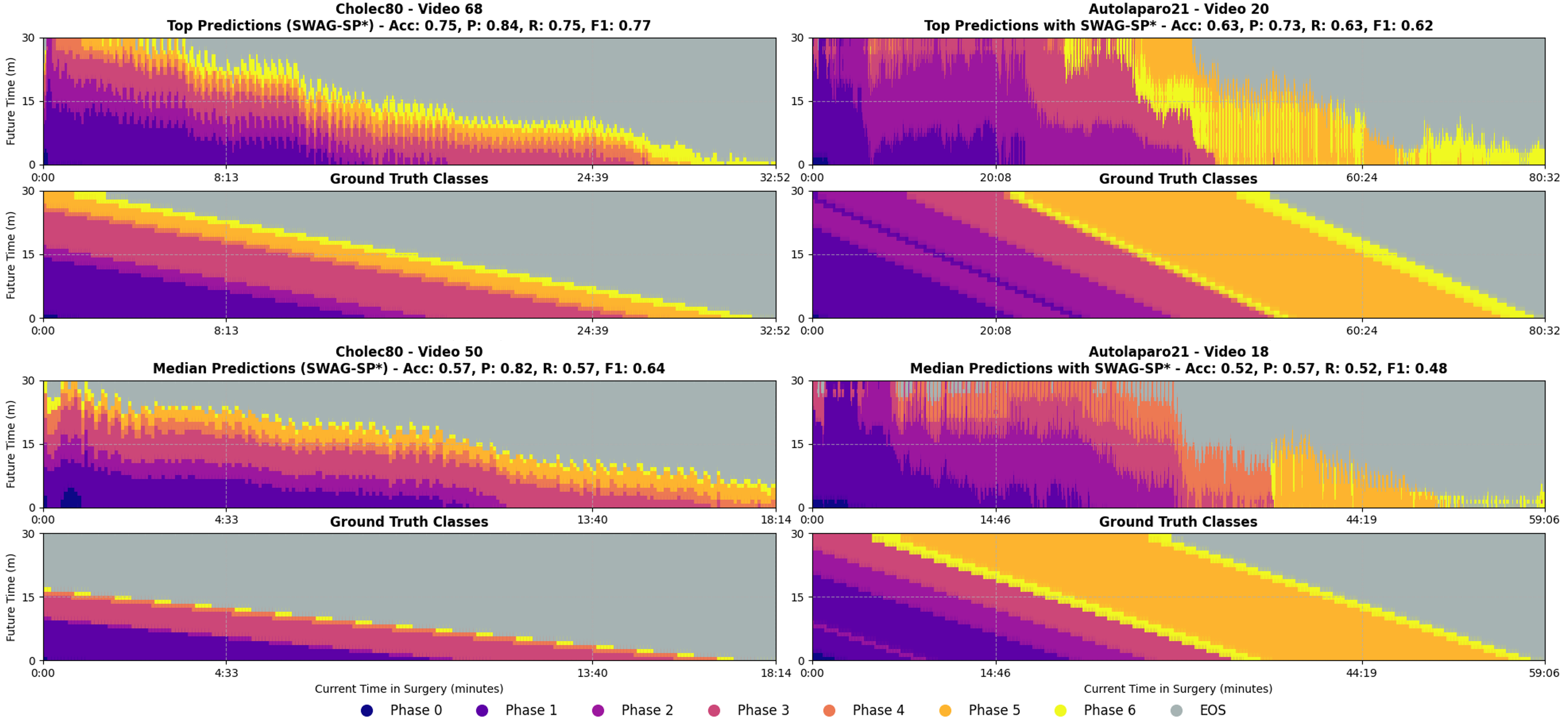}
    \caption{Qualitative results on high- and mid-performing test samples for phase recognition and anticipation using the SWAG-SP* model. Each block represents a single video from either Cholec80 or AutoLaparo21: the top two rows correspond to top-performing cases, and the bottom two to median-performing ones. For each video, the upper panel shows predicted phase segments over a 30-minute anticipation horizon; the lower panel shows the corresponding ground-truth. The $x$-axis indicates the current time within the surgery; the $y$-axis corresponds to the future anticipation horizon. Colours represent surgical phases, with grey denoting the end-of-surgery class.}
    \label{fig:top_mid_performing_videos}
\end{figure}

\begin{figure}[h]
    \includegraphics[width=1.0\linewidth]{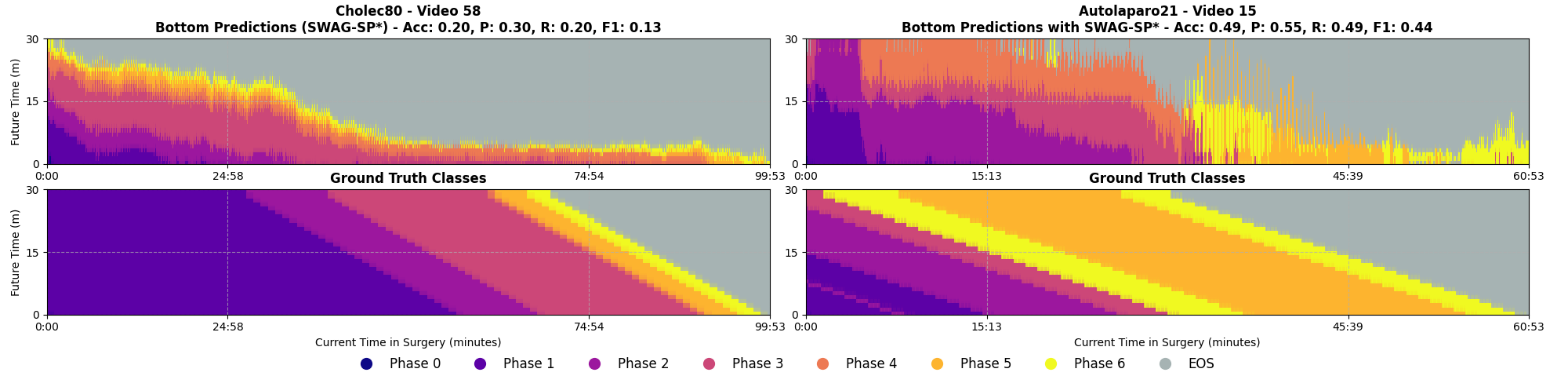}
    \caption{Qualitative results on two bottom-performing test samples using the SWAG-SP* model. Each example shows predicted (top) and ground-truth (bottom) phase segments over a 30-minute anticipation horizon. The $x$-axis represents surgical time progression, and the $y$-axis denotes anticipation into the future. Colours indicate surgical phases; grey corresponds to the end-of-surgery class.}
    \label{fig:bottom_performing_video}
\end{figure}

Figures~\ref{fig:top_mid_performing_videos} and~\ref{fig:bottom_performing_video} provide qualitative comparisons for recognition and anticipation tasks across both datasets. Top-performing cases show predicted phase sequences that closely match ground truth timing and durations, while bottom-performing videos exhibit systematic temporal misalignment where predictions consistently over- or under-estimate phase lengths. The model maintains smooth, continuous predictions across all performance levels.

\subsection{Surgical Phase Anticipation with Remaining Time} Based on the classification evaluation results, we select the single-pass (SWAG-SP) decoder for comparison on state-of-the-art methods for remaining time until the next phase occurrence (see Table~\ref{tab:remaining_time_duration}). For 2-minute and 3-minute horizons, SWAG-SP achieves the best inMAE scores (0.54 and 0.77 minutes for 2 and 3 minutes), outperforming Bayesian~\cite{rivoir2020rethinking} and IIA-Net~\cite{YUAN2022102611}. At the 5-minute horizon, SWAG-SP ranks second with its wMAE score, showing strong adaptability for mid-range anticipations.

\begin{table}[h]
\centering
\caption{\textbf{Remaining time to next phase occurrence at 2, 3, and 5 minutes horizons.} MAEs (in minutes) for \textit{inside} and \textit{outside} the anticipation window, and \textit{weighted} mean. The results from previous methods are reported in IIA-Net~\cite{YUAN2022102611}.
}
\label{tab:remaining_time_duration}
\begin{tabular}{l|ccc|ccc|ccc}
\hline
\multirow{3}{*}{Methods} & \multicolumn{9}{c}{\textbf{Cholec80 (60-20)}} \\
 & \multicolumn{3}{c|}{\textbf{wMAE $[\downarrow]$}} & \multicolumn{3}{c|}{\textbf{inMAE $[\downarrow]$}} & \multicolumn{3}{c}{\textbf{outMAE $[\downarrow]$}} \\
 & \textbf{2 min} & \textbf{3 min} & \textbf{5 min} & \textbf{2 min} & \textbf{3 min} & \textbf{5 min} & \textbf{2 min} & \textbf{3 min} & \textbf{5 min} \\
\hline
Bayesian\cite{rivoir2020rethinking} &  0.39 & 0.59 & 0.85 & 0.63 & 0.86 &  \underline{1.17} & 0.15 & 0.32 & 0.52 \\
IIA-Net~\cite{YUAN2022102611} † & \underline{0.36} & \underline{0.49} & \cellcolor{lightblue!25}\textbf{0.68} &  \underline{0.62} &  \underline{0.81} & \cellcolor{lightblue!25}\textbf{1.08} & \underline{0.10} & \underline{0.18} & \cellcolor{lightblue!25}\textbf{0.28} \\
\hline
SWAG-SP       & \cellcolor{lightblue!25}\textbf{0.32} &  \cellcolor{lightblue!25}\textbf{0.48} &  \underline{0.80} & \cellcolor{lightblue!25}\textbf{0.54} & \cellcolor{lightblue!25}\textbf{0.77} & 1.26 &  \cellcolor{lightblue!25}\textbf{0.09} &  \cellcolor{lightblue!25}\textbf{0.17} &  \underline{0.34} \\
\hline
\end{tabular}
\begin{tablenotes}
\footnotesize
\item \textbf{Note:} The light blue cells in bold and underlined values are the best and second-best results, respectively. † indicates methods using more data annotations for supervision i.e. bounding boxes and segmentation masks.
\end{tablenotes}
\end{table}

IIA-Net~\cite{YUAN2022102611} uses additional manual annotations like instrument bounding boxes and segmentation maps for supervision, while our approach relies solely on phase labels. In Table \ref{tab:rsd-cholec80}, we compare our SWAG-SP method with previous approaches on the Remaining Surgery Duration (RSD) task. Using 4-fold cross-validation, our model ranks second on MAE-5 and MAE-ALL, outperforming previous methods, except BD-Net, all methods exclusively designed for this task.

\begin{table}[h]
\centering
\caption{\textbf{Remaining Surgery Duration (RSD) estimation on Cholec80.} Those results were reported in BD-Net.}
\label{tab:rsd-cholec80}
\footnotesize
\begin{tabular}{l|ccc}
\hline
\textbf{Methods} & \textbf{MAE-5 (min)} & \textbf{MAE-30 (min)} & \textbf{MAE-ALL (min)} \\
\hline
TimeLSTM \cite{aksamentov2017deep} & 3.00 ± 1.87 & \underline{5.30} ± 1.86 & 8.27 ± 6.25 \\
RSDNet \cite{twinanda2018rsdnet} & 8.36 ± 3.71 & 6.83 ± 2.57 & 10.01 ± 6.54 \\
CataNet \cite{marafioti2021catanet} & 2.47 ± 2.62 & 5.53 ± 2.75 & 8.27 ± 6.81 \\
BD-Net \cite{wu2023bdnet}* & \cellcolor{lightblue!25}\textbf{1.97} ± 1.54 & \cellcolor{lightblue!25}\textbf{4.84 ± 2.31} & \cellcolor{lightblue!25}\textbf{7.75 ± 6.43} \\
SWAG-SP (ours) & \underline{2.24} ± 2.32 & 6.48 ± 3.85 & \underline{8.18} ± 5.33 \\
\hline
\end{tabular}
\begin{tablenotes}
\footnotesize
\item \textbf{Note:} * This method randomly created 4-splits for validation, whereas we used consecutive 60/20 splits for training and testing, respectively.
\end{tablenotes}
\end{table}

\begin{figure}[h]
    \centering 
    \includegraphics[width=1.0\linewidth]{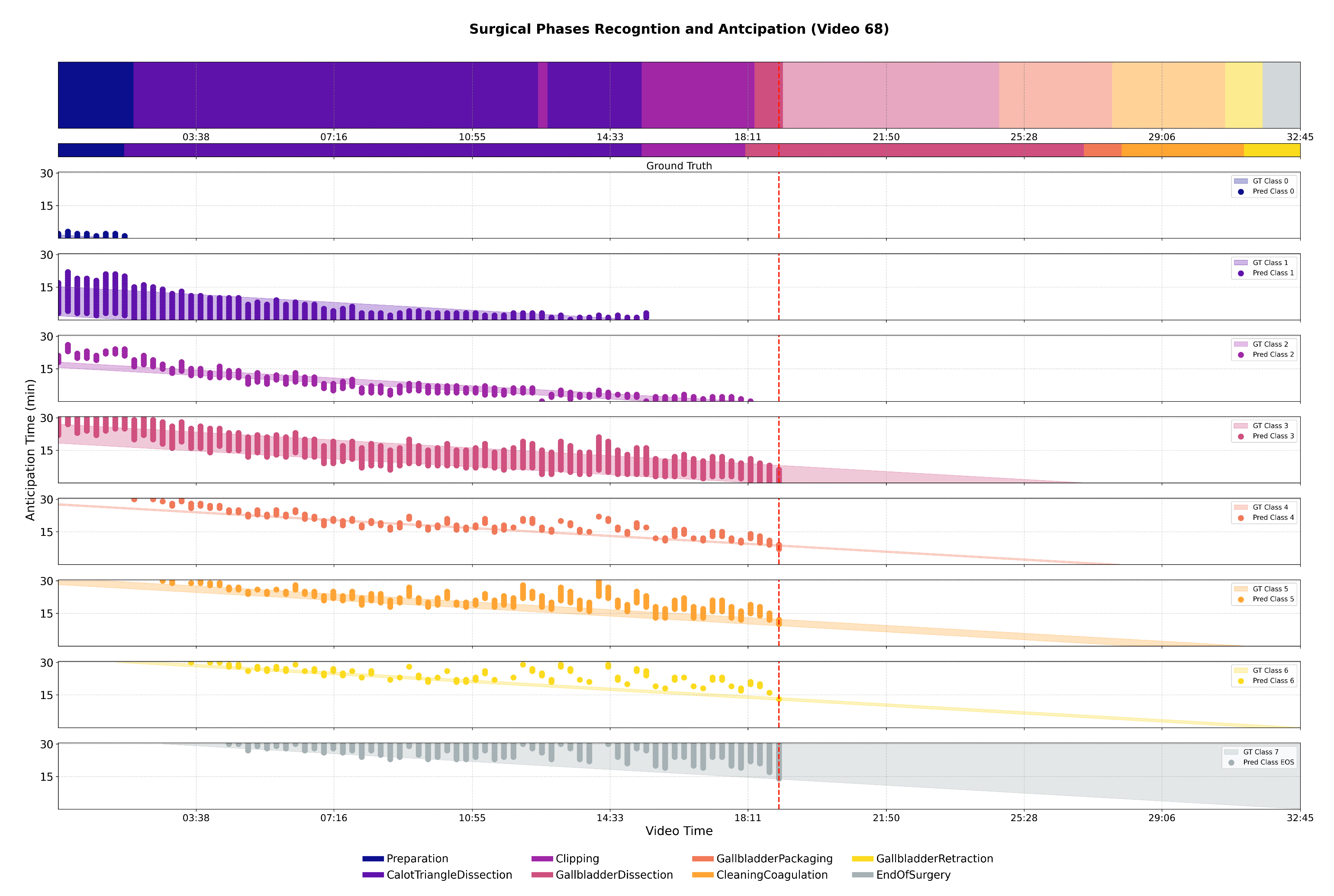}
    \caption{Surgical phase recognition and anticipation within a 30-minute horizon (y-axis) for Cholec80’s video 68, with the current elapsed time (x-axis) indicated by the red dashed vertical line. The thin, second row shows the ground truth phase segments, while the top rectangle displays the classified frames into phases; note that the more transparent colours to the right of the dashed line indicate the generated future phases at 60-second intervals. The last eight rectangles depict the previous remaining time predictions per class mapping to the 1-d timeline at the top.}
    \label{fig:classification_task_h15}
\end{figure}

Fig. ~\ref{fig:classification_task_h15} illustrates how predicted future phase segments would be depicted during inference. We extend the standard visualisation method for the recognition task by completing the missing part on the right side in an online setting with the predicted remaining surgical workflow. Our method proposed to fill this gap in a simple continuous approach using dense temporal segment classification, similarly to the recognition task but with a forward thinking approach. We believe this novel segment-based anticipation visualisation method could improve intraoperative awareness and guidance.

\section{Discussion}\label{sec:discussion}

Our study reformulates surgical phase anticipation as a generative sequence modeling task, introducing SWAG with transformer-based approaches under auto-regressive (AR) and single-pass (SP) decoding strategies, enhanced through token-level conditioning with prior knowledge (SP*).

Our results reveal distinct performance patterns that reflect the underlying complexity of different surgical procedures. On Cholec80, with its highly structured and predictable workflow, the Naive2 baseline achieves remarkably strong anticipation performance (F1 = 39.5\%), demonstrating that simple approaches can excel when procedures follow consistent phase ordering. However, this advantage disappears on the more complex AutoLaparo21 dataset, where our SP* model demonstrates clear superiority (F1 = 41.3\%, SegF1 = 34.8\%), significantly outperforming naive baselines. This contrast highlights a fundamental insight: as surgical complexity increases, sophisticated modeling approaches become essential for accurate anticipation. The differential performance of our approaches provides valuable insights into anticipation strategies. Single-pass decoding with remaining time regression (R2C) achieves stronger performance on structured procedures like Cholec80 (F1 = 36.1\% vs 32.9\% on AutoLaparo21), suggesting that regression-based approaches work well when phase durations are predictable. Conversely, sequence generation approaches (SP*) excel on variable procedures, better capturing the complex temporal dependencies and phase transitions characteristic of less standardised surgeries. The superior segment-level performance of our approaches compared to naive methods (despite similar frame-level scores) demonstrates the importance of maintaining temporal coherence in predictions. We successfully extend anticipation horizons beyond the previous 5-minute limit, with SWAG-SP achieving the lowest wMAE and inMAE for short-term horizons while bridging into remaining surgery duration prediction. However, performance degrades significantly beyond 15-20 minute horizons, with F1 scores dropping below 30\% for most methods, indicating fundamental challenges in long-term surgical workflow prediction even for standardized procedures.

\subsection*{Limitations and Future Directions}
Our work reveals that long-term surgical anticipation remains challenging, with performance rapidly degrading over extended horizons due to the inherent unpredictability of intraoperative events and decision-making. Current limitations include reliance on relatively small datasets and the assumption of single valid future trajectories. Future work should prioritise three key directions: (1) incorporating uncertainty estimation to better handle the stochastic nature of surgical workflows, (2) developing evaluation frameworks that accommodate multiple plausible surgical trajectories rather than assuming deterministic workflows, and (3) scaling to larger, more diverse surgical datasets enriched with fine-grained action annotations and instrument control data. Looking toward longer-term advances, exploring generative approaches to create novel surgical scenarios beyond existing expert demonstrations could train robots to handle unprecedented cases, enhancing the current standard of surgical care and elevating patient safety to unprecedented levels.

\section{Conclusion}\label{sec:conclusion}
This work introduces SWAG, a unified encoder-decoder framework for surgical phase recognition and long-term anticipation, successfully demonstrating the value of generative sequence modeling enhanced with prior knowledge conditioning. Our evaluation across Cholec80 and AutoLaparo21 datasets reveals that while simple approaches suffice for structured procedures, sophisticated modeling becomes essential for complex, variable surgeries. SWAG's superior performance on challenging datasets and its ability to maintain temporal coherence in predictions establishes it as a promising foundation for intraoperative guidance systems, while highlighting the fundamental challenges that remain in long-term surgical workflow anticipation.

\section*{Declarations}
The authors declare no conflict of interest. This article does not contain any studies with human participants or animals performed by any of the authors. This manuscript does not contain any patient data.

\bibliography{sn-article}


\begin{thebibliography}{35}
\ifx \bisbn   \undefined \def \bisbn  #1{ISBN #1}\fi
\ifx \binits  \undefined \def \binits#1{#1}\fi
\ifx \bauthor  \undefined \def \bauthor#1{#1}\fi
\ifx \batitle  \undefined \def \batitle#1{#1}\fi
\ifx \bjtitle  \undefined \def \bjtitle#1{#1}\fi
\ifx \bvolume  \undefined \def \bvolume#1{\textbf{#1}}\fi
\ifx \byear  \undefined \def \byear#1{#1}\fi
\ifx \bissue  \undefined \def \bissue#1{#1}\fi
\ifx \bfpage  \undefined \def \bfpage#1{#1}\fi
\ifx \blpage  \undefined \def \blpage #1{#1}\fi
\ifx \burl  \undefined \def \burl#1{\textsf{#1}}\fi
\ifx \doiurl  \undefined \def \doiurl#1{\url{https://doi.org/#1}}\fi
\ifx \betal  \undefined \def \betal{\textit{et al.}}\fi
\ifx \binstitute  \undefined \def \binstitute#1{#1}\fi
\ifx \binstitutionaled  \undefined \def \binstitutionaled#1{#1}\fi
\ifx \bctitle  \undefined \def \bctitle#1{#1}\fi
\ifx \beditor  \undefined \def \beditor#1{#1}\fi
\ifx \bpublisher  \undefined \def \bpublisher#1{#1}\fi
\ifx \bbtitle  \undefined \def \bbtitle#1{#1}\fi
\ifx \bedition  \undefined \def \bedition#1{#1}\fi
\ifx \bseriesno  \undefined \def \bseriesno#1{#1}\fi
\ifx \blocation  \undefined \def \blocation#1{#1}\fi
\ifx \bsertitle  \undefined \def \bsertitle#1{#1}\fi
\ifx \bsnm \undefined \def \bsnm#1{#1}\fi
\ifx \bsuffix \undefined \def \bsuffix#1{#1}\fi
\ifx \bparticle \undefined \def \bparticle#1{#1}\fi
\ifx \barticle \undefined \def \barticle#1{#1}\fi
\bibcommenthead
\ifx \bconfdate \undefined \def \bconfdate #1{#1}\fi
\ifx \botherref \undefined \def \botherref #1{#1}\fi
\ifx \url \undefined \def \url#1{\textsf{#1}}\fi
\ifx \bchapter \undefined \def \bchapter#1{#1}\fi
\ifx \bbook \undefined \def \bbook#1{#1}\fi
\ifx \bcomment \undefined \def \bcomment#1{#1}\fi
\ifx \oauthor \undefined \def \oauthor#1{#1}\fi
\ifx \citeauthoryear \undefined \def \citeauthoryear#1{#1}\fi
\ifx \endbibitem  \undefined \def \endbibitem {}\fi
\ifx \bconflocation  \undefined \def \bconflocation#1{#1}\fi
\ifx \arxivurl  \undefined \def \arxivurl#1{\textsf{#1}}\fi
\csname PreBibitemsHook\endcsname

\bibitem[\protect\citeauthoryear{Sexton et~al.}{2018}]{Sexton148}
\begin{botherref}
\oauthor{\bsnm{Sexton}, \binits{K.}},
\oauthor{\bsnm{Johnson}, \binits{A.}},
\oauthor{\bsnm{Gotsch}, \binits{A.}},
\oauthor{\bsnm{Hussein}, \binits{A.A.}},
\oauthor{\bsnm{Cavuoto}, \binits{L.}},
\oauthor{\bsnm{Guru}, \binits{K.A.}}:
Anticipation, teamwork and cognitive load: chasing efficiency during robot-assisted surgery.
BMJ Quality \& Safety
(2018)
\end{botherref}
\endbibitem

\bibitem[\protect\citeauthoryear{Yurko et~al.}{2010}]{Yurko2010}
\begin{botherref}
\oauthor{\bsnm{Yurko}, \binits{Y.Y.}},
\oauthor{\bsnm{Scerbo}, \binits{M.W.}},
\oauthor{\bsnm{Prabhu}, \binits{A.S.}},
\oauthor{\bsnm{Acker}, \binits{C.E.}},
\oauthor{\bsnm{Stefanidis}, \binits{D.}}:
Higher mental workload is associated with poorer laparoscopic performance as measured by the nasa-tlx tool.
Simulation in Healthcare
(2010)
\end{botherref}
\endbibitem

\bibitem[\protect\citeauthoryear{Czempiel et~al.}{2020}]{czempiel2020tecno}
\begin{botherref}
\oauthor{\bsnm{Czempiel}, \binits{T.}},
\oauthor{\bsnm{Paschali}, \binits{M.}},
\oauthor{\bsnm{Keicher}, \binits{M.}},
\oauthor{\bsnm{Simson}, \binits{W.}},
\oauthor{\bsnm{Feußner}, \binits{H.}},
\oauthor{\bsnm{Kim}, \binits{S.T.}},
\oauthor{\bsnm{Navab}, \binits{N.}}:
Tecno: Surgical phase recognition with multi-stage temporal convolutional networks.
International Conference on Medical Image Computing and Computer-Assisted Intervention
(2020)
\end{botherref}
\endbibitem

\bibitem[\protect\citeauthoryear{Liu et~al.}{2023a}]{liu2023lovit}
\begin{botherref}
\oauthor{\bsnm{Liu}, \binits{Y.}},
\oauthor{\bsnm{Boels}, \binits{M.}},
\oauthor{\bsnm{García-Peraza-Herrera}, \binits{L.C.}},
\oauthor{\bsnm{Vercauteren}, \binits{T.K.M.}},
\oauthor{\bsnm{Dasgupta}, \binits{P.}},
\oauthor{\bsnm{Granados}, \binits{A.}},
\oauthor{\bsnm{Ourselin}, \binits{S.}}:
Lovit: Long video transformer for surgical phase recognition.
Medical Image Analysis
(2023)
\end{botherref}
\endbibitem

\bibitem[\protect\citeauthoryear{Liu et~al.}{2023b}]{liu2023skit}
\begin{bchapter}
\bauthor{\bsnm{Liu}, \binits{Y.}},
\bauthor{\bsnm{Huo}, \binits{J.}},
\bauthor{\bsnm{Peng}, \binits{J.}},
\bauthor{\bsnm{Sparks}, \binits{R.}},
\bauthor{\bsnm{Dasgupta}, \binits{P.}},
\bauthor{\bsnm{Granados}, \binits{A.}},
\bauthor{\bsnm{Ourselin}, \binits{S.}}:
\bctitle{Skit: a fast key information video transformer for online surgical phase recognition}.
In: \bbtitle{Proceedings of the IEEE/CVF International Conference on Computer Vision}
(\byear{2023})
\end{bchapter}
\endbibitem

\bibitem[\protect\citeauthoryear{Aksamentov et~al.}{2017}]{aksamentov2017deep}
\begin{bchapter}
\bauthor{\bsnm{Aksamentov}, \binits{I.}},
\bauthor{\bsnm{Twinanda}, \binits{A.P.}},
\bauthor{\bsnm{Mutter}, \binits{D.}},
\bauthor{\bsnm{Marescaux}, \binits{J.}},
\bauthor{\bsnm{Padoy}, \binits{N.}}:
\bctitle{Deep neural networks predict remaining surgery duration from cholecystectomy videos}.
In: \bbtitle{Medical Image Computing and Computer-Assisted Intervention − MICCAI 2017},
pp. \bfpage{586}--\blpage{593}.
\bpublisher{Springer},
\blocation{Cham}
(\byear{2017})
\end{bchapter}
\endbibitem

\bibitem[\protect\citeauthoryear{Twinanda et~al.}{2018}]{twinanda2018rsdnet}
\begin{botherref}
\oauthor{\bsnm{Twinanda}, \binits{A.P.}},
\oauthor{\bsnm{Yengera}, \binits{G.}},
\oauthor{\bsnm{Mutter}, \binits{D.}},
\oauthor{\bsnm{Marescaux}, \binits{J.}},
\oauthor{\bsnm{Padoy}, \binits{N.}}:
Rsdnet: Learning to predict remaining surgery duration from laparoscopic videos without manual annotations.
IEEE transactions on medical imaging
(2018)
\end{botherref}
\endbibitem

\bibitem[\protect\citeauthoryear{Rivoir et~al.}{2019}]{rivoir2019unsupervised}
\begin{bchapter}
\bauthor{\bsnm{Rivoir}, \binits{D.}},
\bauthor{\bsnm{Bodenstedt}, \binits{S.}},
\bauthor{\bsnm{Bechtolsheim}, \binits{F.}},
\bauthor{\bsnm{Distler}, \binits{M.}},
\bauthor{\bsnm{Weitz}, \binits{J.}},
\bauthor{\bsnm{Speidel}, \binits{S.}}:
\bctitle{Unsupervised temporal video segmentation as an auxiliary task for predicting the remaining surgery duration}.
In: \bbtitle{OR 2.0 Context-Aware Operating Theaters and Machine Learning in Clinical Neuroimaging},
pp. \bfpage{29}--\blpage{37}.
\bpublisher{Springer},
\blocation{Cham}
(\byear{2019})
\end{bchapter}
\endbibitem

\bibitem[\protect\citeauthoryear{Marafioti et~al.}{2021}]{marafioti2021catanet}
\begin{bchapter}
\bauthor{\bsnm{Marafioti}, \binits{A.}},
\bauthor{\bsnm{Hayoz}, \binits{M.}},
\bauthor{\bsnm{Gallardo}, \binits{M.}},
\bauthor{\bsnm{M{\'a}rquez~Neila}, \binits{P.}},
\bauthor{\bsnm{Wolf}, \binits{S.}},
\bauthor{\bsnm{Zinkernagel}, \binits{M.}},
\bauthor{\bsnm{Sznitman}, \binits{R.}}:
\bctitle{Catanet: Predicting remaining cataract surgery duration}.
In: \bbtitle{Medical Image Computing and Computer Assisted Intervention -- MICCAI 2021},
pp. \bfpage{426}--\blpage{435}.
\bpublisher{Springer},
\blocation{Cham}
(\byear{2021})
\end{bchapter}
\endbibitem

\bibitem[\protect\citeauthoryear{Wu et~al.}{2023}]{wu2023bdnet}
\begin{botherref}
\oauthor{\bsnm{Wu}, \binits{J.}},
\oauthor{\bsnm{Zou}, \binits{X.}},
\oauthor{\bsnm{Tao}, \binits{R.}},
\oauthor{\bsnm{Zheng}, \binits{G.}}:
Nonlinear regression of remaining surgery duration from videos via bayesian lstm-based deep negative correlation learning.
Computerized Medical Imaging and Graphics
(2023)
\end{botherref}
\endbibitem

\bibitem[\protect\citeauthoryear{Wijekoon et~al.}{2024}]{wijekoon2024pitrsdnet}
\begin{botherref}
\oauthor{\bsnm{Wijekoon}, \binits{A.}},
\oauthor{\bsnm{Das}, \binits{A.}},
\oauthor{\bsnm{Herrera}, \binits{R.R.}},
\oauthor{\bsnm{Khan}, \binits{D.Z.}},
\oauthor{\bsnm{Hanrahan}, \binits{J.}},
\oauthor{\bsnm{Carter}, \binits{E.}},
\oauthor{\bsnm{Luoma}, \binits{V.}},
\oauthor{\bsnm{Stoyanov}, \binits{D.}},
\oauthor{\bsnm{Marcus}, \binits{H.J.}},
\oauthor{\bsnm{Bano}, \binits{S.}}:
Pitrsdnet: Predicting intra-operative remaining surgery duration in endoscopic pituitary surgery.
arXiv preprint arXiv: 2409.16998
(2024)
\end{botherref}
\endbibitem

\bibitem[\protect\citeauthoryear{Rivoir et~al.}{2020}]{rivoir2020rethinking}
\begin{botherref}
\oauthor{\bsnm{Rivoir}, \binits{D.}},
\oauthor{\bsnm{Bodenstedt}, \binits{S.}},
\oauthor{\bsnm{Funke}, \binits{I.}},
\oauthor{\bsnm{Bechtolsheim}, \binits{F.}},
\oauthor{\bsnm{Distler}, \binits{M.}},
\oauthor{\bsnm{Weitz}, \binits{J.}},
\oauthor{\bsnm{Speidel}, \binits{S.}}:
Rethinking anticipation tasks: Uncertainty-aware anticipation of sparse surgical instrument usage for context-aware assistance.
International Conference on Medical Image Computing and Computer-Assisted Intervention
(2020)
\end{botherref}
\endbibitem

\bibitem[\protect\citeauthoryear{Yuan et~al.}{2022}]{YUAN2022102611}
\begin{botherref}
\oauthor{\bsnm{Yuan}, \binits{K.}},
\oauthor{\bsnm{Holden}, \binits{M.}},
\oauthor{\bsnm{Gao}, \binits{S.}},
\oauthor{\bsnm{Lee}, \binits{W.}}:
Anticipation for surgical workflow through instrument interaction and recognized signals.
Medical Image Analysis
(2022)
\end{botherref}
\endbibitem

\bibitem[\protect\citeauthoryear{Radford et~al.}{2019}]{radford2019language}
\begin{botherref}
\oauthor{\bsnm{Radford}, \binits{A.}},
\oauthor{\bsnm{Wu}, \binits{J.}},
\oauthor{\bsnm{Child}, \binits{R.}},
\oauthor{\bsnm{Luan}, \binits{D.}},
\oauthor{\bsnm{Amodei}, \binits{D.}},
\oauthor{\bsnm{Sutskever}, \binits{I.}}:
Language models are unsupervised multitask learners.
OpenAI blog
(2019)
\end{botherref}
\endbibitem

\bibitem[\protect\citeauthoryear{Wang et~al.}{2023}]{wang2023memoryandanticipation}
\begin{barticle}
\bauthor{\bsnm{Wang}, \binits{J.}},
\bauthor{\bsnm{Chen}, \binits{G.}},
\bauthor{\bsnm{Huang}, \binits{Y.}},
\bauthor{\bsnm{Wang}, \binits{L.}},
\bauthor{\bsnm{Lu}, \binits{T.}}:
\batitle{Memory-and-anticipation transformer for online action understanding}.
\bjtitle{IEEE International Conference on Computer Vision}
(\byear{2023})
\doiurl{10.1109/ICCV51070.2023.01271}
\end{barticle}
\endbibitem

\bibitem[\protect\citeauthoryear{Blum et~al.}{2010}]{DBLP:conf/miccai/BlumFN10}
\begin{bchapter}
\bauthor{\bsnm{Blum}, \binits{T.}},
\bauthor{\bsnm{Feu{\ss}ner}, \binits{H.}},
\bauthor{\bsnm{Navab}, \binits{N.}}:
\bctitle{Modeling and segmentation of surgical workflow from laparoscopic video}.
In: \bbtitle{Medical Image Computing and Computer-Assisted Intervention -- MICCAI 2010},
pp. \bfpage{400}--\blpage{407}.
\bpublisher{Springer},
\blocation{Berlin, Heidelberg}
(\byear{2010})
\end{bchapter}
\endbibitem

\bibitem[\protect\citeauthoryear{Twinanda}{2017}]{DBLP:phd/hal/Twinanda17}
\begin{botherref}
\oauthor{\bsnm{Twinanda}, \binits{A.P.}}:
Vision-based approaches for surgical activity recognition using laparoscopic and {RBGD} videos.
PhD thesis,
University of Strasbourg
(2017)
\end{botherref}
\endbibitem

\bibitem[\protect\citeauthoryear{Vaswani et~al.}{2017}]{vaswani2017attention}
\begin{botherref}
\oauthor{\bsnm{Vaswani}, \binits{A.}},
\oauthor{\bsnm{Shazeer}, \binits{N.M.}},
\oauthor{\bsnm{Parmar}, \binits{N.}},
\oauthor{\bsnm{Uszkoreit}, \binits{J.}},
\oauthor{\bsnm{Jones}, \binits{L.}},
\oauthor{\bsnm{Gomez}, \binits{A.N.}},
\oauthor{\bsnm{Kaiser}, \binits{L.}},
\oauthor{\bsnm{Polosukhin}, \binits{I.}}:
Attention is all you need.
Neural Information Processing Systems
(2017)
\end{botherref}
\endbibitem

\bibitem[\protect\citeauthoryear{Dosovitskiy et~al.}{2020}]{dosovitskiy2020image}
\begin{botherref}
\oauthor{\bsnm{Dosovitskiy}, \binits{A.}},
\oauthor{\bsnm{Beyer}, \binits{L.}},
\oauthor{\bsnm{Kolesnikov}, \binits{A.}},
\oauthor{\bsnm{Weissenborn}, \binits{D.}},
\oauthor{\bsnm{Zhai}, \binits{X.}},
\oauthor{\bsnm{Unterthiner}, \binits{T.}},
\oauthor{\bsnm{Dehghani}, \binits{M.}},
\oauthor{\bsnm{Minderer}, \binits{M.}},
\oauthor{\bsnm{Heigold}, \binits{G.}},
\oauthor{\bsnm{Gelly}, \binits{S.}},
\oauthor{\bsnm{Uszkoreit}, \binits{J.}},
\oauthor{\bsnm{Houlsby}, \binits{N.}}:
An image is worth 16x16 words: Transformers for image recognition at scale.
International Conference on Learning Representations
(2020)
\end{botherref}
\endbibitem

\bibitem[\protect\citeauthoryear{Gao et~al.}{2021}]{gao2021transsvnet}
\begin{bchapter}
\bauthor{\bsnm{Gao}, \binits{X.}},
\bauthor{\bsnm{Jin}, \binits{Y.}},
\bauthor{\bsnm{Long}, \binits{Y.}},
\bauthor{\bsnm{Dou}, \binits{Q.}},
\bauthor{\bsnm{Heng}, \binits{P.}}:
\bctitle{Trans-svnet: Accurate phase recognition from surgical videos via hybrid embedding aggregation transformer}.
In: \beditor{\bsnm{Bruijne}, \binits{M.}},
\beditor{\bsnm{Cattin}, \binits{P.C.}},
\beditor{\bsnm{Cotin}, \binits{S.}},
\beditor{\bsnm{Padoy}, \binits{N.}},
\beditor{\bsnm{Speidel}, \binits{S.}},
\beditor{\bsnm{Zheng}, \binits{Y.}},
\beditor{\bsnm{Essert}, \binits{C.}} (eds.)
\bbtitle{Medical Image Computing and Computer Assisted Intervention - {MICCAI} 2021}.
\bpublisher{Springer}, \blocation{???}
(\byear{2021})
\end{bchapter}
\endbibitem

\bibitem[\protect\citeauthoryear{Ding and Li}{2022}]{ding2022exploring}
\begin{barticle}
\bauthor{\bsnm{Ding}, \binits{X.}},
\bauthor{\bsnm{Li}, \binits{X.}}:
\batitle{Exploring segment-level semantics for online phase recognition from surgical videos}.
\bjtitle{{IEEE} Trans. Medical Imaging}
\bvolume{41}(\bissue{11}),
\bfpage{3309}--\blpage{3319}
(\byear{2022})
\end{barticle}
\endbibitem

\bibitem[\protect\citeauthoryear{Rivoir et~al.}{2024}]{rivoir2024pitfalls}
\begin{barticle}
\bauthor{\bsnm{Rivoir}, \binits{D.}},
\bauthor{\bsnm{Funke}, \binits{I.}},
\bauthor{\bsnm{Speidel}, \binits{S.}}:
\batitle{On the pitfalls of batch normalization for end-to-end video learning: A study on surgical workflow analysis}.
\bjtitle{Medical Image Analysis}
\bvolume{94},
\bfpage{103126}
(\byear{2024})
\doiurl{10.1016/j.media.2024.103126}
\end{barticle}
\endbibitem

\bibitem[\protect\citeauthoryear{Yang et~al.}{2024}]{yang2024surgformer}
\begin{barticle}
\bauthor{\bsnm{Yang}, \binits{S.}},
\bauthor{\bsnm{Luo}, \binits{L.}},
\bauthor{\bsnm{Wang}, \binits{Q.}},
\bauthor{\bsnm{Chen}, \binits{H.}}:
\batitle{Surgformer: Surgical transformer with hierarchical temporal attention for surgical phase recognition}.
\bjtitle{International Conference on Medical Image Computing and Computer-Assisted Intervention}
(\byear{2024})
\doiurl{10.48550/arXiv.2408.03867}
\end{barticle}
\endbibitem

\bibitem[\protect\citeauthoryear{Ban et~al.}{2021}]{ban2021suprgan}
\begin{barticle}
\bauthor{\bsnm{Ban}, \binits{Y.}},
\bauthor{\bsnm{Rosman}, \binits{G.}},
\bauthor{\bsnm{Eckhoff}, \binits{J.}},
\bauthor{\bsnm{Ward}, \binits{T.M.}},
\bauthor{\bsnm{Hashimoto}, \binits{D.}},
\bauthor{\bsnm{Kondo}, \binits{T.}},
\bauthor{\bsnm{Iwaki}, \binits{H.}},
\bauthor{\bsnm{Meireles}, \binits{O.}},
\bauthor{\bsnm{Rus}, \binits{D.}}:
\batitle{Supr-gan: Surgical prediction gan for event anticipation in laparoscopic and robotic surgery}.
\bjtitle{IEEE Robotics and Automation Letters}
(\byear{2021})
\doiurl{10.1109/lra.2022.3156856}
\end{barticle}
\endbibitem

\bibitem[\protect\citeauthoryear{Boels et~al.}{2024}]{boels2024supra}
\begin{botherref}
\oauthor{\bsnm{Boels}, \binits{M.}},
\oauthor{\bsnm{Liu}, \binits{Y.}},
\oauthor{\bsnm{Dasgupta}, \binits{P.}},
\oauthor{\bsnm{Granados}, \binits{A.}},
\oauthor{\bsnm{Ourselin}, \binits{S.}}:
Supra: Surgical phase recognition and anticipation for intra-operative planning.
arXiv preprint arXiv: 2403.06200
(2024)
\end{botherref}
\endbibitem

\bibitem[\protect\citeauthoryear{Zhang et~al.}{2022}]{zhang2022towards}
\begin{barticle}
\bauthor{\bsnm{Zhang}, \binits{X.}},
\bauthor{\bsnm{Moubayed}, \binits{N.A.}},
\bauthor{\bsnm{Shum}, \binits{H.P.H.}}:
\batitle{Towards graph representation learning based surgical workflow anticipation}.
\bjtitle{IEEE-EMBS International Conference on Biomedical and Health Informatics (BHI)}
(\byear{2022})
\doiurl{10.1109/BHI56158.2022.9926801}
\end{barticle}
\endbibitem

\bibitem[\protect\citeauthoryear{Yin et~al.}{2024}]{yin2024hypergraphtransformer}
\begin{botherref}
\oauthor{\bsnm{Yin}, \binits{L.}},
\oauthor{\bsnm{Ban}, \binits{Y.}},
\oauthor{\bsnm{Eckhoff}, \binits{J.}},
\oauthor{\bsnm{Meireles}, \binits{O.}},
\oauthor{\bsnm{Rus}, \binits{D.}},
\oauthor{\bsnm{Rosman}, \binits{G.}}:
Hypergraph-transformer (hgt) for interactive event prediction in laparoscopic and robotic surgery.
arXiv preprint arXiv: 2402.01974
(2024)
\end{botherref}
\endbibitem

\bibitem[\protect\citeauthoryear{Ginesi et~al.}{2020}]{ginesi2020autonomous}
\begin{botherref}
\oauthor{\bsnm{Ginesi}, \binits{M.}},
\oauthor{\bsnm{Meli}, \binits{D.}},
\oauthor{\bsnm{Roberti}, \binits{A.}},
\oauthor{\bsnm{Sansonetto}, \binits{N.}},
\oauthor{\bsnm{Fiorini}, \binits{P.}}:
Autonomous task planning and situation awareness in robotic surgery.
IEEE/RJS International Conference on Intelligent RObots and Systems
(2020)
\end{botherref}
\endbibitem

\bibitem[\protect\citeauthoryear{Qin et~al.}{2020}]{qin2020davincinet}
\begin{botherref}
\oauthor{\bsnm{Qin}, \binits{Y.}},
\oauthor{\bsnm{Feyzabadi}, \binits{S.}},
\oauthor{\bsnm{Allan}, \binits{M.}},
\oauthor{\bsnm{Burdick}, \binits{J.}},
\oauthor{\bsnm{Azizian}, \binits{M.}}:
davincinet: Joint prediction of motion and surgical state in robot-assisted surgery.
IEEE/RJS International Conference on Intelligent RObots and Systems
(2020)
\end{botherref}
\endbibitem

\bibitem[\protect\citeauthoryear{Zhang et~al.}{2023}]{zhang2023laparoscopic}
\begin{botherref}
\oauthor{\bsnm{Zhang}, \binits{J.}},
\oauthor{\bsnm{Zhou}, \binits{S.}},
\oauthor{\bsnm{Wang}, \binits{Y.}},
\oauthor{\bsnm{Shi}, \binits{S.}},
\oauthor{\bsnm{Wan}, \binits{C.}},
\oauthor{\bsnm{Zhao}, \binits{H.}},
\oauthor{\bsnm{Cai}, \binits{X.}},
\oauthor{\bsnm{Ding}, \binits{H.}}:
Laparoscopic image-based critical action recognition and anticipation with explainable features.
IEEE Journal of Biomedical and Health Informatics
(2023)
\end{botherref}
\endbibitem

\bibitem[\protect\citeauthoryear{Girdhar and Grauman}{2021}]{girdhar2021anticipative}
\begin{barticle}
\bauthor{\bsnm{Girdhar}, \binits{R.}},
\bauthor{\bsnm{Grauman}, \binits{K.}}:
\batitle{Anticipative video transformer}.
\bjtitle{IEEE International Conference on Computer Vision}
(\byear{2021})
\doiurl{10.1109/ICCV48922.2021.01325}
\end{barticle}
\endbibitem

\bibitem[\protect\citeauthoryear{Twinanda et~al.}{2017}]{twinanda2017endonet}
\begin{barticle}
\bauthor{\bsnm{Twinanda}, \binits{A.P.}},
\bauthor{\bsnm{Shehata}, \binits{S.}},
\bauthor{\bsnm{Mutter}, \binits{D.}},
\bauthor{\bsnm{Marescaux}, \binits{J.}},
\bauthor{\bsnm{Mathelin}, \binits{M.}},
\bauthor{\bsnm{Padoy}, \binits{N.}}:
\batitle{Endonet: {A} deep architecture for recognition tasks on laparoscopic videos}.
\bjtitle{{IEEE} Trans. Medical Imaging}
\bvolume{36}(\bissue{1}),
\bfpage{86}--\blpage{97}
(\byear{2017})
\end{barticle}
\endbibitem

\bibitem[\protect\citeauthoryear{Wang et~al.}{2022}]{wang2022autolaparo}
\begin{bchapter}
\bauthor{\bsnm{Wang}, \binits{Z.}},
\bauthor{\bsnm{Lu}, \binits{B.}},
\bauthor{\bsnm{Long}, \binits{Y.}},
\bauthor{\bsnm{Zhong}, \binits{F.}},
\bauthor{\bsnm{Cheung}, \binits{T.-H.}},
\bauthor{\bsnm{Dou}, \binits{Q.}},
\bauthor{\bsnm{Liu}, \binits{Y.}}:
\bctitle{Autolaparo: A new dataset of integrated multi-tasks for image-guided surgical automation in laparoscopic hysterectomy}.
In: \bbtitle{International Conference on Medical Image Computing and Computer-Assisted Intervention},
pp. \bfpage{486}--\blpage{496}
(\byear{2022}).
\bcomment{Springer}
\end{bchapter}
\endbibitem

\bibitem[\protect\citeauthoryear{Funke et~al.}{2025}]{funke2025tunes}
\begin{botherref}
\oauthor{\bsnm{Funke}, \binits{I.}},
\oauthor{\bsnm{Rivoir}, \binits{D.}},
\oauthor{\bsnm{Krell}, \binits{S.}},
\oauthor{\bsnm{Speidel}, \binits{S.}}:
Tunes: A temporal u-net with self-attention for video-based surgical phase recognition.
IEEE Transactions on Biomedical Engineering
(2025)
\end{botherref}
\endbibitem

\bibitem[\protect\citeauthoryear{Jin et~al.}{2022}]{jin2022trans}
\begin{botherref}
\oauthor{\bsnm{Jin}, \binits{Y.}},
\oauthor{\bsnm{Long}, \binits{Y.}},
\oauthor{\bsnm{Gao}, \binits{X.}},
\oauthor{\bsnm{Stoyanov}, \binits{D.}},
\oauthor{\bsnm{Dou}, \binits{Q.}},
\oauthor{\bsnm{Heng}, \binits{P.-A.}}:
Trans-svnet: hybrid embedding aggregation transformer for surgical workflow analysis.
International Journal of Computer Assisted Radiology and Surgery
(2022)
\end{botherref}
\endbibitem

\end{thebibliography}

\newpage  

\section{Supplementary Material}\label{supp-sec:supp-material}

\subsection{Implementation Details}\label{supp-sec:supp-details}
Our experiments were conducted on a single NVIDIA Tesla V100 GPU. We used a 12-head, 12-layer Transformer encoder as our spatial feature extractor, based on the ViT-B/16 architecture following LoViT. This model was pre-trained on ImageNet 1K (IN1k) and produced 768-d representations, with an input image size of 248$\times$248 pixels. For training the spatial feature extractor, we used stochastic gradient descent with momentum for 35 epochs, with a 5-epoch warm-up period and a 30-epoch cosine annealed decay. We used a batch size of 16 and a learning rate of 0.1, which was multiplied by 0.1 at the 20th and 30th epochs. We set the weight decay to 1e-4 and the momentum to 0.9. For the Temporal Self-Attention, we used an input clip length $l$ of 1440 frames or 24 minutes at 1fps with a sliding context length window $w$ of 20 frames, generating 512-d feature vectors. Key-pooled feature dimensions $d$ are 64-d and 32-d on Cholec80 and AutoLaparo21, respectively. The temporal modules underwent training for 40 epochs using SGD and momentum with a learning rate of 3e-4, weight decay of 1e-5, a 5 epoch warm-up period, and a 35 epoch cosine annealed decay, with a batch size of 8.

\subsection{Future Tokens Embedding Initialization}\label{supp-sec:token-embedding}

This process involves iterating over all future token indices \( h_n \) to generate the probability vector \( \mathbf{p}_t \). We sample from the transition probability matrix \( \mathbf{P} \) using the current class index \( i \), which corresponds to the ground-truth label when using teacher forcing during training, or the model's predicted class otherwise. The index \( h_n \) represents the position of the future token within the anticipation horizon \( h_N \). We define the transition probability of being in a future class \( j \) given the current class \( i \) as follows:
\begin{equation}
    p(y_{t+h_n\cdot60} = j \mid y_t = i) = \mathbf{P}[i, j, h_n]
\end{equation}
with 
\begin{equation}
    \quad i \in \{0, \dots, C-1\}, \quad j \in \{0, \dots, C\}, \quad h_n \in \{h_1, \dots, h_N\}
\end{equation}
where \( C \) represents the total number of surgical phases, and \( C \) corresponds to the additional end-of-sequence (EOS) class included for padding purposes.
We define the probability vector \( \mathbf{p}_t \) as a collection of class probability vectors for each future horizon \( h_n \):
\begin{equation}
    \mathbf{p}_t = \bigl[\mathbf{P}[i, :, h_n]\bigr]_{n=1}^{N}, \quad h_n \in \{h_1, \dots, h_N\}
\end{equation}
where \( \mathbf{P}[i, :, h_n] \) corresponds to the probability vector for the future token at index \( h_n \), sampled from the \( i \)-th row of the matrix \( \mathbf{P} \), which is based on the current class index.
Each future token embedding \( \mathbf{q}_t \) is computed by combining an embedding \( u_t \), initialized using the Xavier uniform distribution, with the linearly transformed higher-dimensional probability vector \( \mathbf{p'}_t \) and a sinusoidal positional encoding. The final embedding serves as the input to the transformer decoder:
\begin{equation}
    \mathbf{p'}_t = W_p \mathbf{p}_t + \text{bias}_p
\end{equation}
\begin{equation}
    \mathbf{q}_t = \text{LayerNorm}(u_t + \alpha \mathbf{p'}_t + \text{PositionalEncoding}(t))
\end{equation}

\subsection{Additional Results}

\begin{figure}[h]
    \centering \includegraphics[width=1.\textwidth]{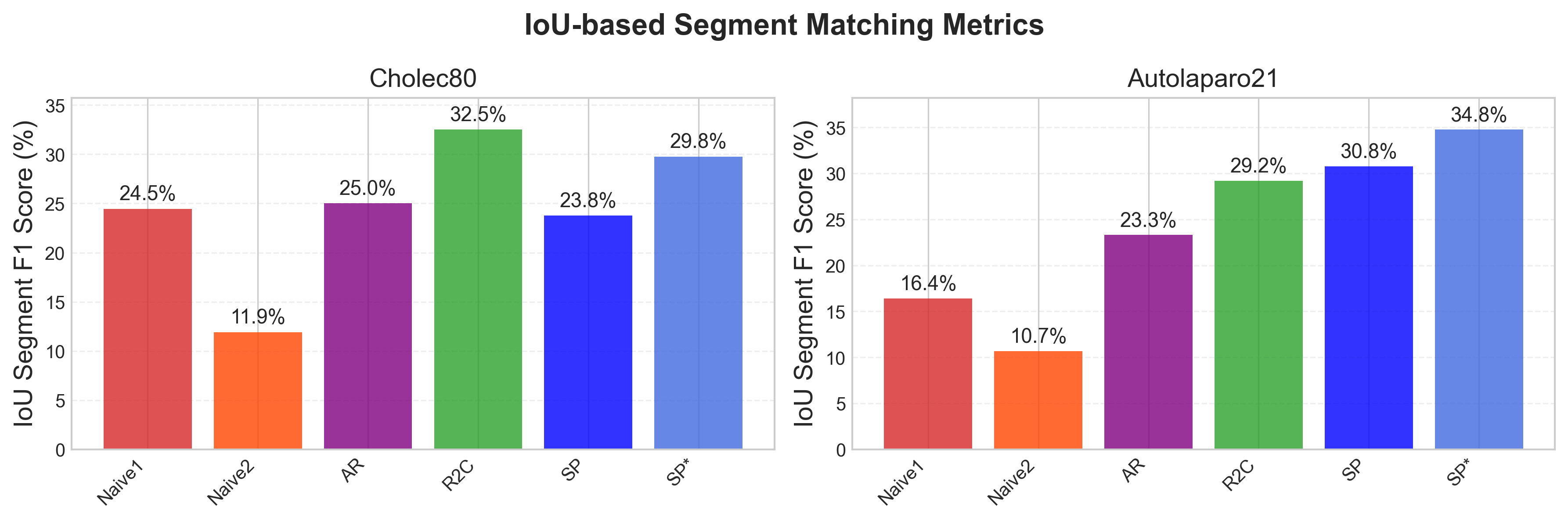}
    \caption{Methods IoU performance for surgical phase recognition and anticipation on Cholec80 and AutoLaparo21.}
    \label{fig:iou}
\end{figure}

\begin{figure}[h]
    \centering \includegraphics[width=1.\textwidth]{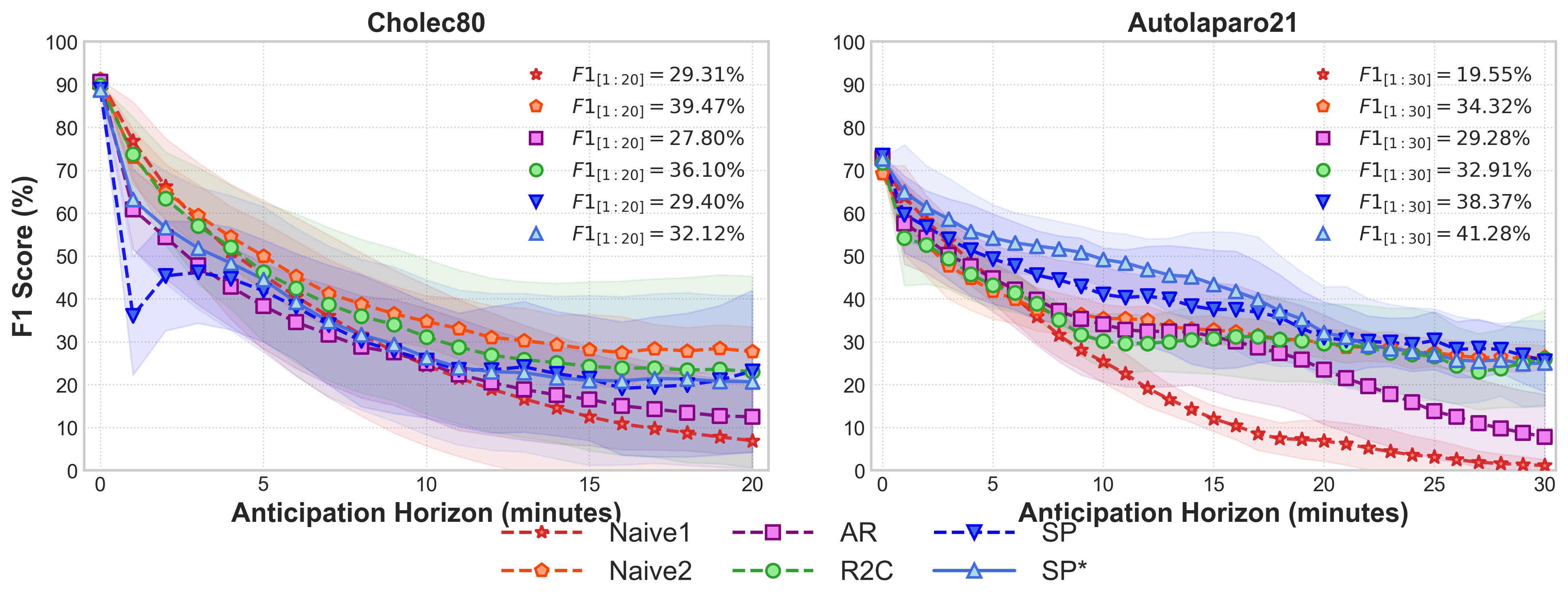}
    \caption{Methods performance for surgical phase recognition and anticipation on Cholec80 and AutoLaparo21. We report the frame-level accuracy scores $Acc_{[1:h_n]}$ and show the mean anticipation scores over $h_n$ minutes (top right corners).}
    \label{fig:acc_scores_over_time}
\end{figure}

\section{Segment-based F1 Score (SegF1) Calculation}

The SegF1 metric evaluates temporal coherence in surgical phase predictions by matching continuous segments rather than individual frames, addressing the problem of oversegmentation.

\subsection{Methodology}

Given a current time point \( t \), our model anticipates future class labels at 1-minute intervals across a fixed anticipation horizon of \( N \) minutes. Let:

\begin{itemize}
    \item \( \hat{y}_{t + h_n \cdot 60} \) denote the predicted label at minute index \( h_n \in \{1, \dots, N\} \)
    \item \( y_{t + h_n \cdot 60} \) denote the corresponding ground-truth label
\end{itemize}

We define the sequences over the horizon as:
\[
\hat{\mathbf{y_t}} = \left( \hat{y}_{t + 60}, \hat{y}_{t + 2 \cdot 60}, \dots, \hat{y}_{t + N \cdot 60} \right), \quad 
\mathbf{y_t} = \left( y_{t + 60}, y_{t + 2 \cdot 60}, \dots, y_{t + N \cdot 60} \right)
\]

\subsubsection{Segment Identification}
From these sequences, we identify continuous segments of the same class. A segment $S = (start, end, c)$ represents consecutive minutes with the same class label $c$, where $start$ and $end$ are minute indices.

The set of ground truth segments $\mathcal{S}^{GT}$ is derived from $\mathbf{y_t}$, and the set of predicted segments $\mathcal{S}^{PR}$ is derived from $\hat{\mathbf{y_t}}$.

\subsubsection{Intersection over Union (IoU) Calculation}
For two segments $S_i = (start_i, end_i, c_i)$ and $S_j = (start_j, end_j, c_j)$, the Intersection over Union (IoU) is:

\begin{align}
\textrm{IoU}(S_i, S_j) &= \frac{|\textrm{Intersection}(S_i, S_j)|}{|\textrm{Union}(S_i, S_j)|} \\
&= \frac{\max(0, \min(end_i, end_j) - \max(start_i, start_j) + 1)}{(end_i - start_i + 1) + (end_j - start_j + 1) - |\textrm{Intersection}(S_i, S_j)|}
\end{align}

\subsubsection{Optimal Matching with Hungarian Algorithm}
A valid match between predicted segment $\hat{S}_i \in \mathcal{S}^{PR}$ and ground truth segment $S_j \in \mathcal{S}^{GT}$ requires:
\[
\hat{c}_i = c_j \quad \text{and} \quad \textrm{IoU}(\hat{S}_i, S_j) \geq \tau
\]
where $\tau = 0.25$ in our implementation, requiring at least 25\% overlap.

We construct a cost matrix $C$ where $C_{i,j} = -\textrm{IoU}(\hat{S}_i, S_j)$ if $\hat{c}_i = c_j$, otherwise $C_{i,j} = \infty$.
Using the Hungarian algorithm, we find the optimal matching $\mathcal{M}$ between predicted and ground truth segments.

\subsubsection{Metric Calculation}
Based on the matching results:
\[
\mathrm{TP} = |\mathcal{M}|, \quad
\mathrm{FP} = |\mathcal{S}^{PR}| - |\mathcal{M}|, \quad
\mathrm{FN} = |\mathcal{S}^{GT}| - |\mathcal{M}|
\]

The SegF1 is then calculated as:
\begin{align}
\mathrm{Precision} &= \frac{\mathrm{TP}}{\mathrm{TP} + \mathrm{FP}} \\
\mathrm{Recall} &= \frac{\mathrm{TP}}{\mathrm{TP} + \mathrm{FN}} \\
\mathrm{SegF1} &= \frac{2 \cdot \mathrm{Precision} \cdot \mathrm{Recall}}{\mathrm{Precision} + \mathrm{Recall}}
\end{align}

\subsection{EOS Class Handling}

For the End-of-Surgery (EOS) class, we apply special handling to prevent its dominance:

\begin{itemize}
\item We limit the number of EOS frames considered per sequence to $\ell_{EoS} = 4$ minutes for Cholec80 and $\ell_{EoS} = 8$ minutes for AutoLaparo21
\item We apply a weighting factor of 0.5 to EOS segments in the calculation of TP, FP, and FN
\end{itemize}

Implementation parameters:
\begin{itemize}
    \item IoU threshold ($\tau$): 0.25
    \item EOS class weighting: 0.5
    \item End-of-Surgery consideration length ($\ell_{EoS}$): 4 minutes (Cholec80), 8 minutes (AutoLaparo21)
    \item Anticipation horizon: 20 minutes (Cholec80), 30 minutes (AutoLaparo21)
\end{itemize}

\subsection{Ablation Studies}\label{supp-sec:ablations}
We conducted ablation experiments to study the impact of various factors on model performance, including context length, anticipation interval, number of context tokens, temporal pooling methods, and model size.

\begin{figure}[h]
\centering
\includegraphics[width=1.0\textwidth]{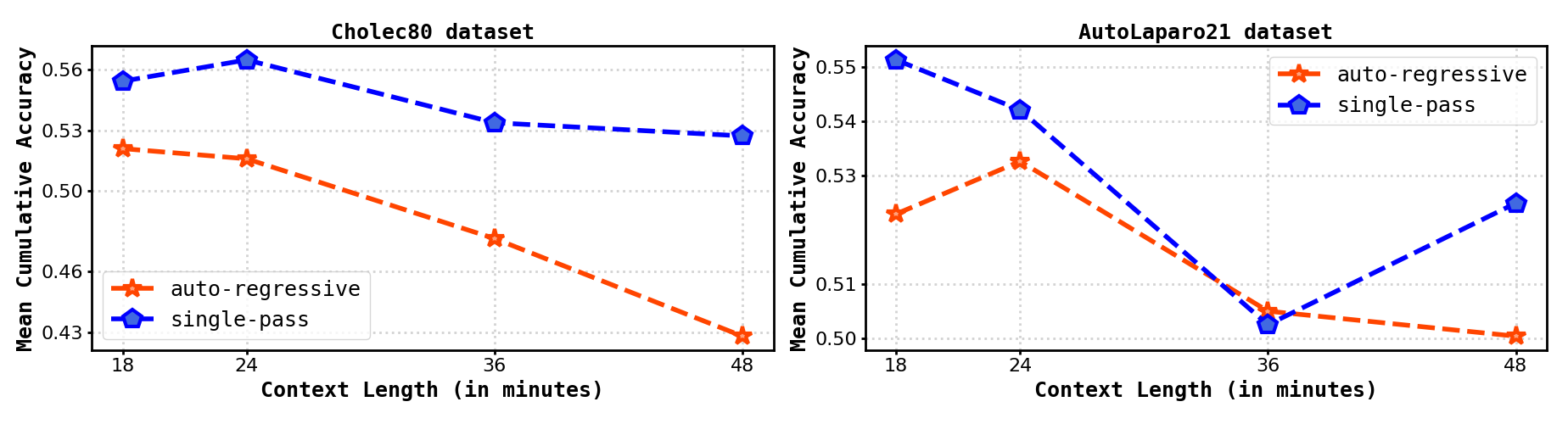}
\caption{Context Length (in minutes) of the input sequence.}
\label{supp-fig:context-length}
\end{figure}

\noindent\textbf{Context Length.} Figure~\ref{supp-fig:context-length} shows that a context length of 24 minutes yields the highest mean cumulative accuracy on both datasets.

\begin{figure}[h]
\centering
\includegraphics[width=1.0\textwidth]{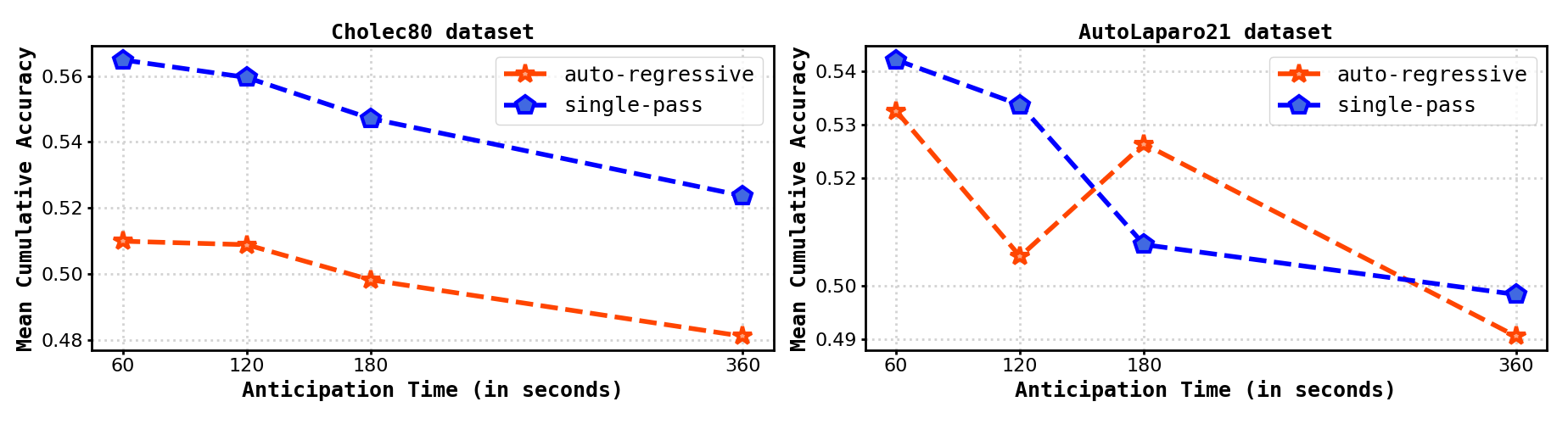}
\caption{Effect of the anticipation time between frames during training (in seconds) on the test accuracy.}
\label{supp-fig:anticipation-time}
\end{figure}

\noindent\textbf{Compression and Anticipation Time.} Figure~\ref{supp-fig:anticipation-time} indicates that a 1-minute interval between input samples produces the highest accuracies on both datasets.

\begin{figure}[h]
\centering
\includegraphics[width=1.0\textwidth]{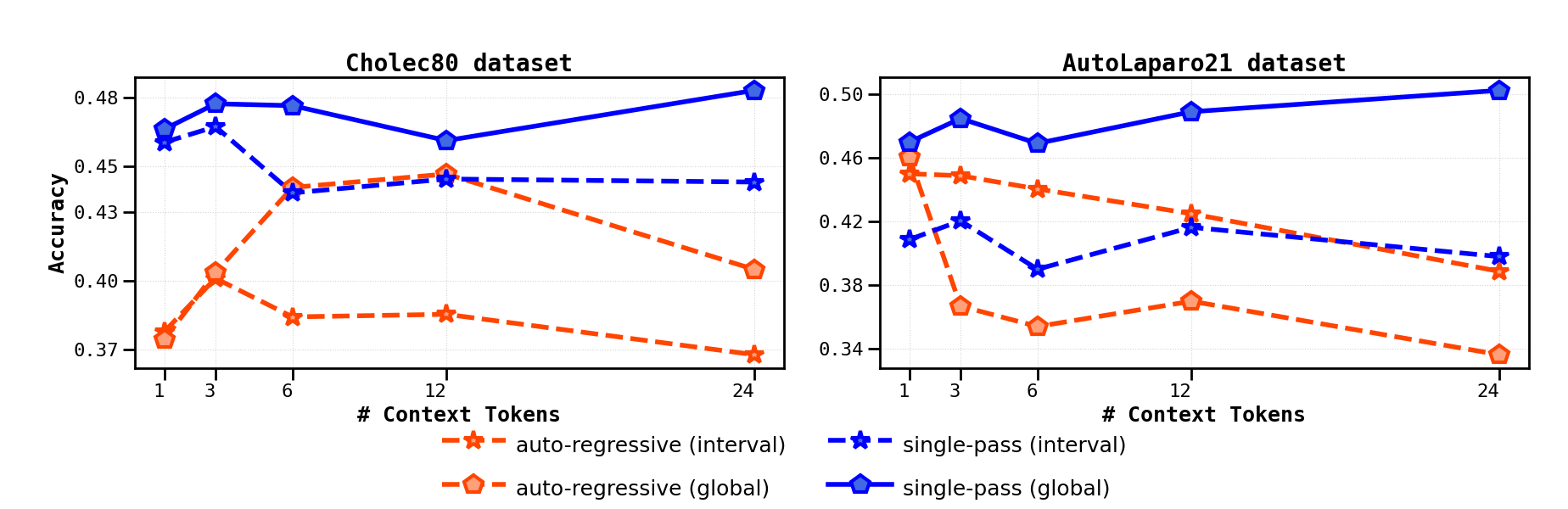}
\caption{Number of context tokens with a fixed context length of 24 minutes i.e. compression of input sequence with global and interval temporal pooling methods.}
\label{supp-fig:num-tokens}
\end{figure}

\noindent\textbf{Temporal Pooling Methods.} Figure~\ref{supp-fig:num-tokens} compares global and interval pooling methods. Single-pass decoding with global context tokens consistently outperforms auto-regressive decoding with either temporal pooling methods for all numbers of context tokens and on both datasets.

\begin{figure}[h]
\centering
\includegraphics[width=1.0\textwidth]{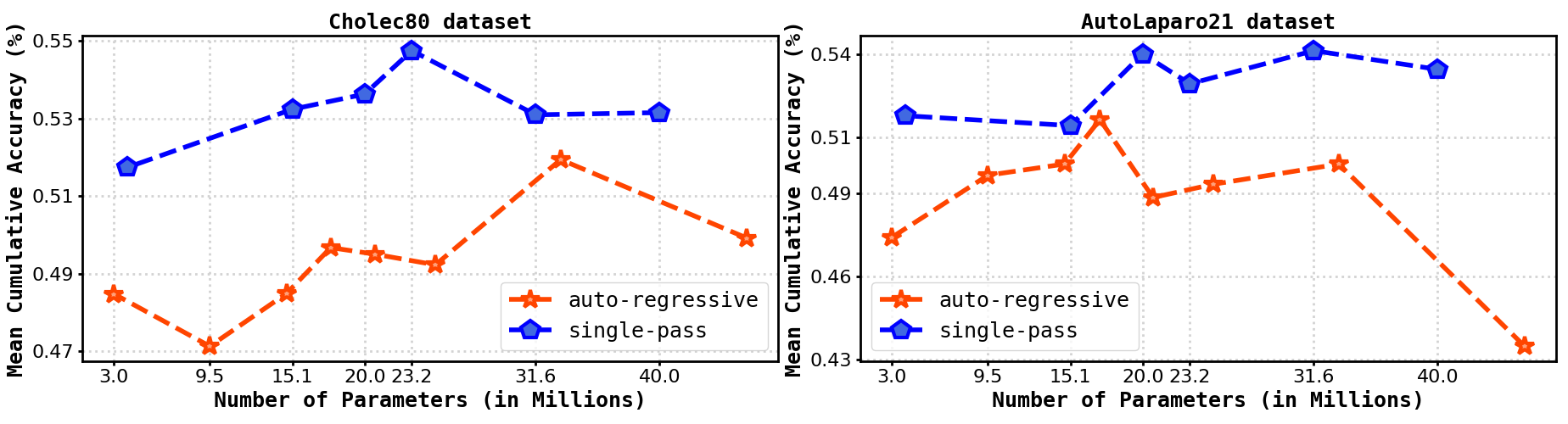}
\caption{Model size (million parameters) over accuracy.}\label{supp-fig:model-size}
\end{figure}

\noindent\textbf{Model Size.} Figure~\ref{supp-fig:model-size} demonstrates that medium-sized models achieve the best overall accuracies, with the optimal model size varying slightly between datasets. These ablations indicate that careful tuning of these parameters can significantly impact long-term surgical phase anticipation performance.

\noindent\textbf{Inside horizon MAE per class.} On the Cholec80 dataset, we observe an increasing error from approximately 2 to 8 minutes for mid- to late-stage surgical phases, specifically classes 4, 5, 6, and EOS. In contrast, anticipation errors for phases 2 and 3 remain stable over time, averaging around 5.5 minutes. The model exhibits an error below 1 minute for phase 1, likely because phase 1 occurs at the very beginning of the surgery, leading the model to predict low values that are often accurate.

\begin{figure}[h]
\centering
\includegraphics[width=1.08\textwidth]{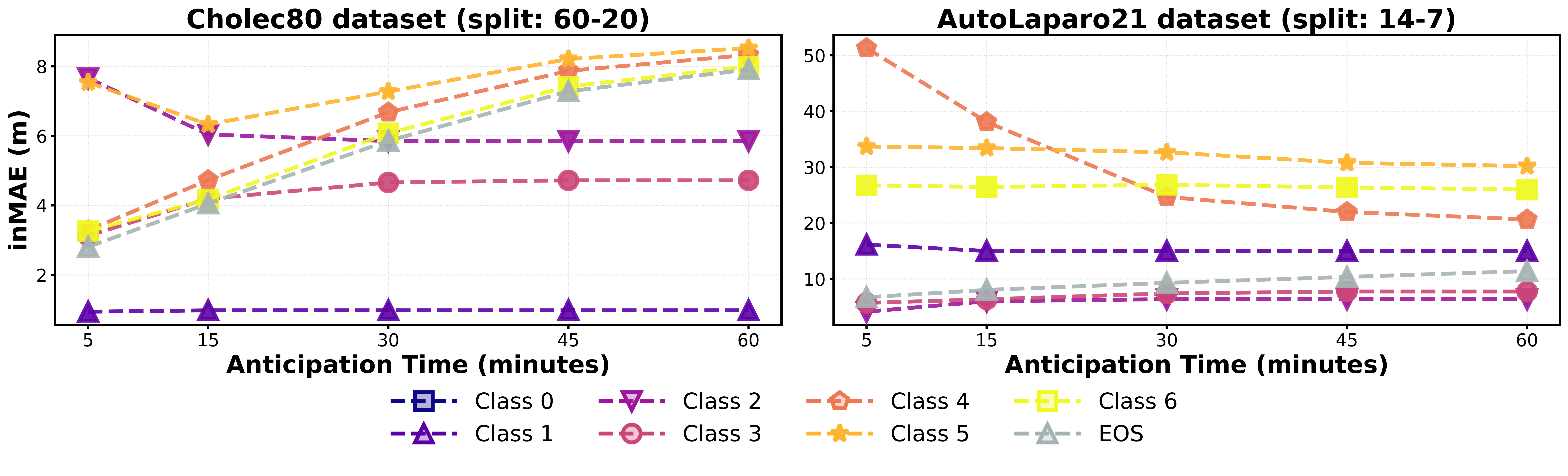}
\caption{MAEs inside the anticipation horizons per-class on Cholec80 and AutoLaparo21.}
\label{supp-fig:inmae-per-class}
\end{figure}

For the AutoLaparo21 dataset, the task is more challenging, especially for mid- to late-phase stages, though the EOS class is less impacted. Errors are less correlated with time, reflecting the intrinsic uncertainty of phase variability. Classes 4, 5, and 6, in particular, are inconsistently present and often exhibit abrupt transitions, making prediction difficult. Over long horizons, the model takes a conservative approach by predicting mid-range values. However, as it nears the 5-minute horizon, it attempts to predict lower, more specific values, which can lead to substantial errors if the phase is absent or has sudden transitions.

\subsection{Our SWAG Application}

We show how a pre-trained language model can translate the predictions into language. Available on our project homepage: \url{https://maxboels.github.io/swag}

\begin{figure}[h]%
\centering
\includegraphics[width=1.\textwidth]{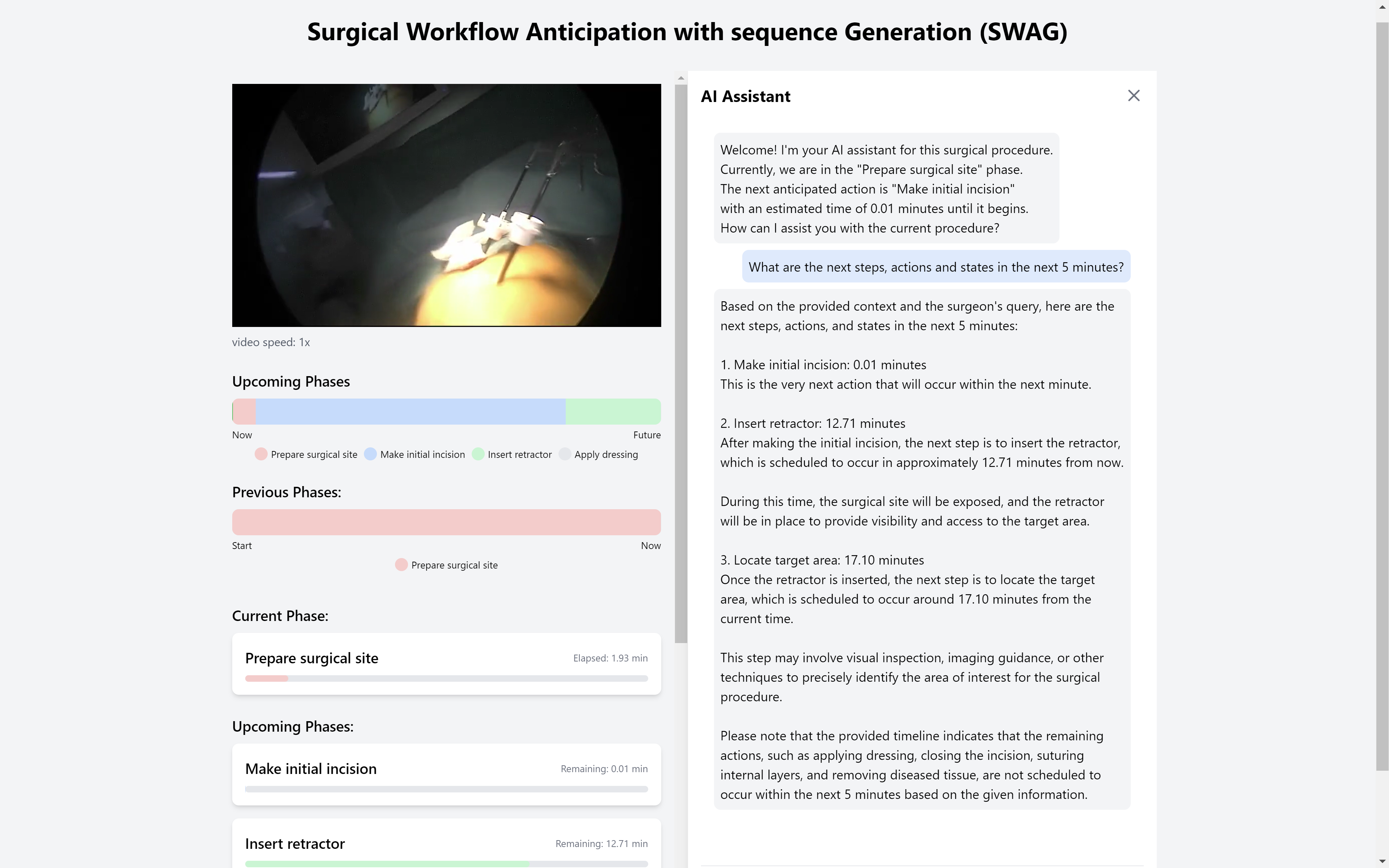}
\caption{SWAG + LLM application.} 
\label{supp-fig:vision-language-app}
\end{figure}

\subsection{Additional Qualitative Results}

\begin{figure}[h!]
\centering
\includegraphics[width=1.\textwidth]{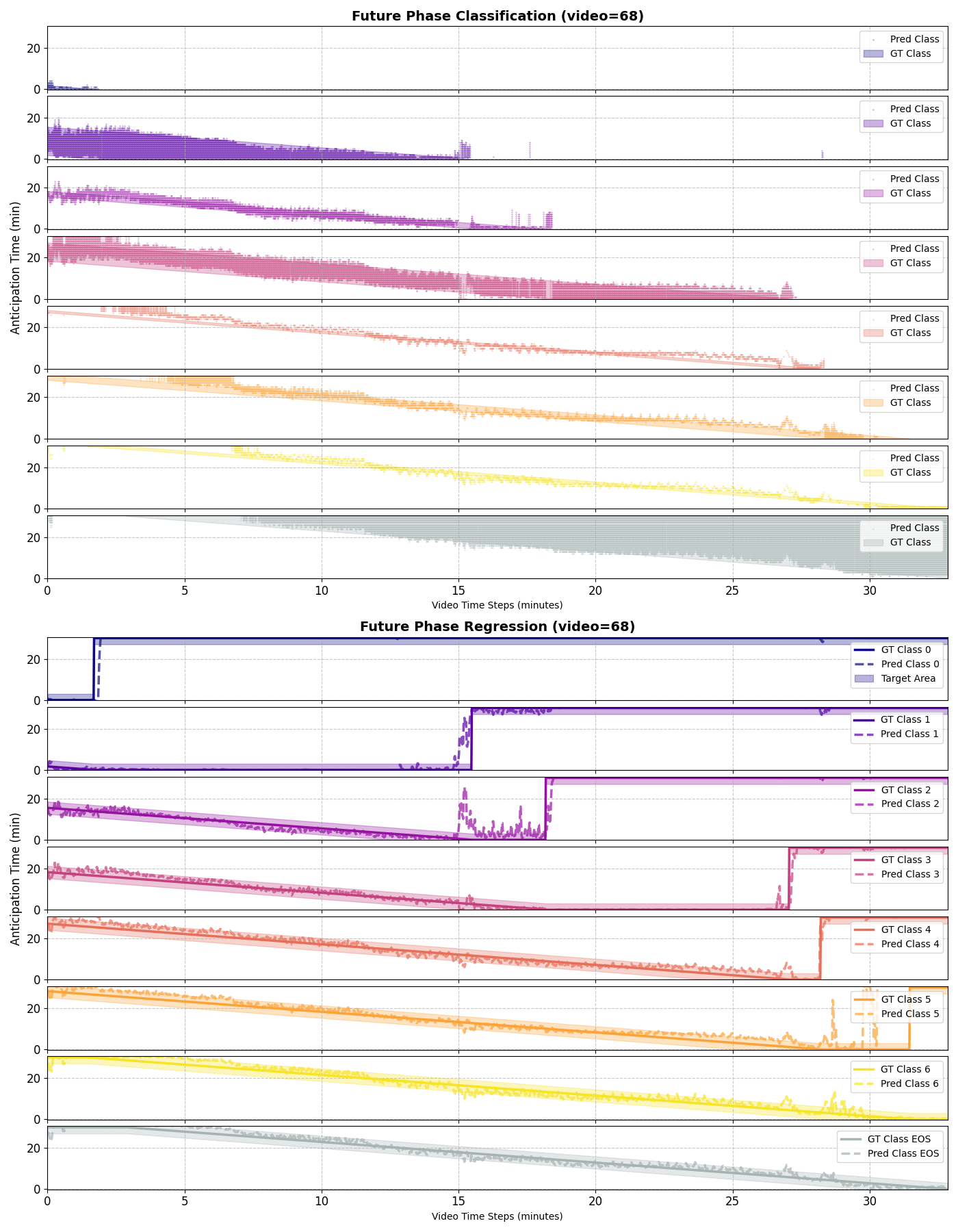}
\caption{Comparison of future phases classification (top) and regression (bottom) task on video 68 from the cholec80 dataset.}
\label{supp-fig:classification-task_h15}
\end{figure}

\end{document}